\documentclass[10pt,twocolumn,letterpaper]{article}

\usepackage{iccv}
\usepackage{times}
\usepackage{epsfig}
\usepackage{graphicx}
\usepackage{amsmath}
\usepackage{amssymb}
\usepackage{pifont}\usepackage{booktabs}
\usepackage{caption}
\usepackage{subcaption}
\usepackage[dvipsnames]{xcolor}
\usepackage[symbol]{footmisc}
\usepackage{CJKutf8}

\usepackage[breaklinks=true,bookmarks=false]{hyperref}

\iccvfinalcopy 

\def\iccvPaperID{11018} 

\ificcvfinal\fi

\definecolor{pretrainingclr}{RGB}{210, 70, 50}
\definecolor{interpolationsclr}{RGB}{30, 120, 50}
\definecolor{vaeclr}{RGB}{34, 62, 191}
\definecolor{gapclr}{RGB}{66, 135, 245}

\newcommand{\cmark}{\color{OliveGreen}{\ding{51}}}\newcommand{\xmark}{\color{Maroon}{\ding{55}}}

\begin{document}

\title{How to Boost Face Recognition with StyleGAN?}

\author{Artem Sevastopolsky$^1$ \qquad
Yury Malkov$^{2,*}$ \qquad
Nikita Durasov$^3$ \\
Luisa Verdoliva$^{1,4}$ \qquad
Matthias Nießner$^1$ \\
\ \\
$^1$ Technical University of Munich, Germany \qquad
$^2$ Twitter, US \\
$^3$ École polytechnique fédérale de Lausanne, Switzerland \\
$^4$ University Federico II of Naples, Italy
}

\twocolumn[{
    \renewcommand\twocolumn[1][]{#1}\maketitle
    \thispagestyle{empty}
    \vspace{-1.0cm}
\begin{center}
        \centering
\includegraphics[trim=3.7cm 5cm 3cm 2.5cm,clip,width=.7\linewidth]{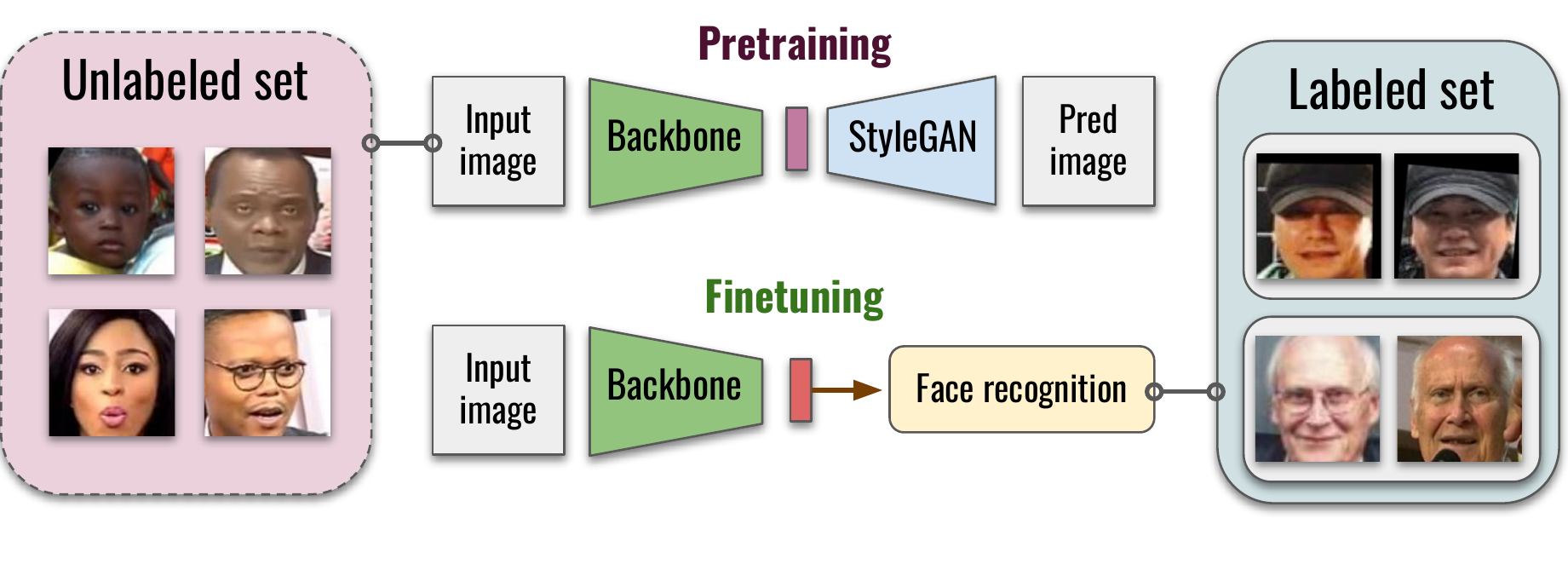}
        \vspace{-0.55cm}
\captionof{figure}{Our method is aimed at boosting the performance of face recognition. This is achieved by gathering a random image collection without face recognition labels (\emph{unlabeled set}) and then fitting a mapping from an image to the StyleGAN latent space onto that collection. To learn this mapping, we use the pSp encoder architecture. For the downstream face recognition task, the same encoder is then fine-tuned on a (potentially much smaller) face recognition dataset with identity labels (\emph{labeled set}).}
        \label{fig:teaser}
    \end{center}}]

\begin{abstract}

\vspace{-0.4cm}

State-of-the-art face recognition systems require vast amounts of labeled training data. 
Given the priority of privacy in face recognition applications, the data is limited to celebrity web crawls, which have issues such as limited numbers of identities.
On the other hand, self-supervised revolution in the industry motivates research on the adaptation of related techniques to facial recognition. 
One of the most popular practical tricks is to augment the dataset by the samples drawn from generative models while preserving the identity. 
We show that a simple approach based on fine-tuning pSp encoder for StyleGAN allows to improve upon the state-of-the-art facial recognition and performs better compared to training on synthetic face identities. 
We also collect large-scale unlabeled datasets with controllable ethnic constitution -- AfricanFaceSet-5M (5 million images of different people) and AsianFaceSet-3M (3 million images of different people) --
and we show that pretraining on each of them improves recognition of the respective ethnicities (as well as others), while combining all unlabeled datasets results in the biggest performance increase.
Our self-supervised strategy is the most useful with limited amounts of labeled training data, which can be beneficial for more tailored face recognition tasks and when facing privacy concerns. 
Evaluation is based on a standard RFW dataset and a new large-scale \linebreak RB-\nolinebreak WebFace benchmark.
The code and data are made publicly available at~\url{https://github.com/seva100/stylegan-for-facerec}.

\end{abstract} \vspace{-0.4cm}
\section{Introduction}
\label{sec:intro}

\begin{figure*}[t!]
    \begin{center}
        \centering
\includegraphics[trim=0.1cm 8.5cm 3.3cm 0cm,clip,width=1.0\textwidth]{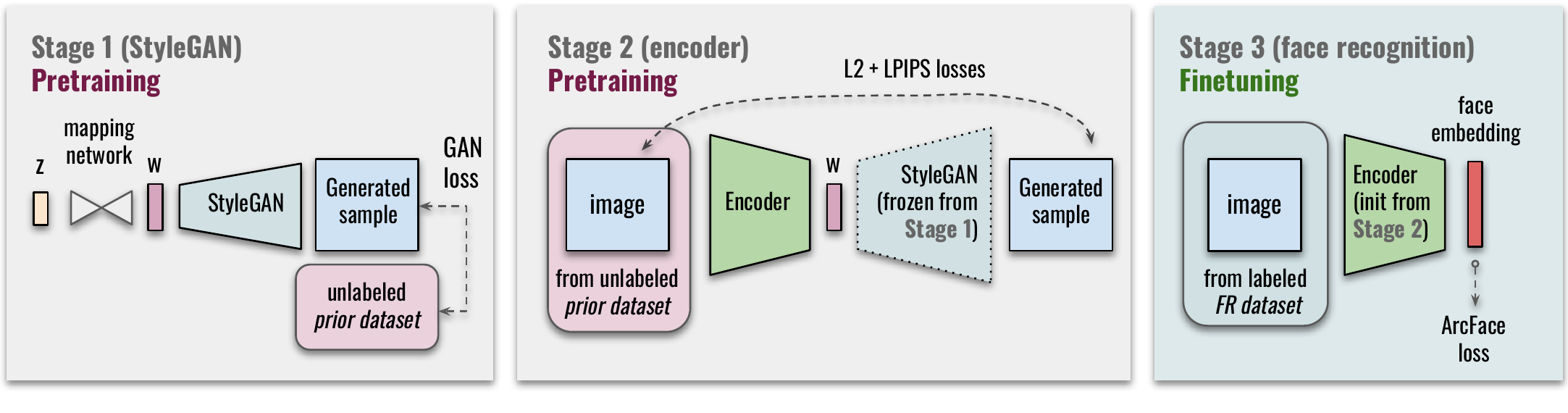}
    \end{center}
    \vspace{-0.5cm}
    \caption{Our method is trained in three consecutive steps. First, we fit StyleGAN2-ADA to the face image distribution of the unlabeled prior dataset $\mathcal{D}^{prior}$. Second, the pSp encoder is trained (also on $\mathcal{D}^{prior}$) to map images to the latent codes in the learned latent space.  Finally, the encoder, which is pretrained to extract meaningful features from an image, is fine-tuned for the downstream face recognition task with the ArcFace loss (similar losses can be used instead) on $\mathcal{D}^{facerec}$. The two first steps comprise the self-supervised pretraining stage; i.e., no identity labels are required for them. }
    \label{fig:method_scheme}
    \vspace{-0.3cm}
\end{figure*}

Modern face recognition methods rely on deep convolutional networks trained on large-scale datasets~\cite{Yi2014learning,Cao2018Vggface2,Guo2016celeb,Zhu2021webface}. These methods are now being integrated into a vast number of real-world applications, ranging from face unlock for smartphones and photo organizers to law enforcement systems and border control. A typical open face recognition dataset consists of web-crawled images of celebrities, leading to limited size and lack of balance in subgroups, such as ethnicity, age, etc. Training a state-of-the-art solution, however, requires enormous amounts of labeled data, scraping which may lead to privacy and legal issues. We suggest and study an alternative solution to using celebrity photos --  pretraining the face recognition backbone on a generative task. Specifically, we first train StyleGAN2-ADA~\cite{Karras2020training} on collected \linebreak unlabeled data (which we later refer to as an \textit{unlabeled prior dataset}) to fit the face image distribution. Subsequently, we train an encoder (following pixel2style2pixel (pSp) architecture~\cite{Richardson2021encoding}) that maps input images to vectors in the learned StyleGAN2-ADA latent space. Importantly, during the pretraining steps, no identity labels are used, so we can use diverse datasets crawled from the Internet without compromising privacy. Finally, we transfer the learned pSp encoder convolutional weights into the face recognition network and train it in a standard face recognition setup.

We show that, in contrast to training face recognition tasks on StyleGAN generated data (also demonstrated in~\cite{nguyen2020recognizing,qiu2021synface} and studied e.g. in~\cite{maluleke2022studying}), our encoder pretraining step significantly boosts the final performance. The idea of augmenting face recognition datasets with synthetic data is widespread and constitutes many approaches, however, unclear and heuristic definition of the target label limits the amount of useful signal that can be transferred into the face recognition model this way. Our approach goes hand-in-hand with the current development of self-supervised learning~\cite{Devlin2018bert,Liu2019roberta,Conneau2019unsupervised,Dosovitskiy2020image} and makes up one of the first approaches of its application to face recognition~\cite{he2022enhancing,lin2021domain}. This allows us to demonstrate vast improvements on limited labeled training data compared to the setup without self-supervised pretraining (for instance, 10\% verification accuracy increase for only 1\% of the labeled data used).

The simplicity of the data collection procedure also allows us to control the distribution of the unlabeled data and thus influence the decrease of the error rates for specific demographic groups. Despite the fact that the current state-of-the-art algorithms often demonstrate very low average error rates~\cite{Deng2019arcface,Wang2018cosface,Wang2021face,Wang2021facex}, 
it is considered unethical to integrate face recognition solutions that exhibit significant ethnic, age, or gender bias. Such bias is present both in open-source face recognition methods~\cite{Gong2021mitigating,Xu2021consistent,Wang2020mitigating,Gong2019debface,Kang2017incorporating,Shi2019probabilistic,wang2023mixfairface} and in comprehensively evaluated commercial face recognition systems~\cite{Grother2019Face}, resulting in significantly different error rates measured for the groups of interest. The topic has attracted significant attention in other areas of computer vision operating in face domain~\cite{Chen2021understanding,Denton2019image,Google2022pixel,Rathgeb2021demographic} and in the media.

We constructively demonstrate that collecting large amounts of in-the-wild face images of a given group of interest is feasible (and can be done semi-automatically), while collecting datasets with identity labels is problematic. The labels require linking photos of the same person taken in different conditions, which means the person must be tracked.  This typically constrains public datasets to celebrities, gathered using search engines~\cite{Guo2016celeb,Yi2014learning,Cao2018Vggface2,Zhu2021webface}, while social networks and companies that provide services with photos use input from users.  Second, the collected in-the-wild data, treated as a set of faces without identity labels (but labeled with the group attribution), can be efficiently used for self-supervised pretraining for face recognition networks, subsequently fine-tuned on the standard face recognition datasets (see Fig.~\ref{fig:teaser}). 

To summarize our main contribution, we present a novel self-supervised method for improving the performance of face recognition based on StyleGAN pretraining. This allows to leverage large-scale amounts of available unlabeled data for face recognition. While the improvement is the most significant on limited data, pretraining is also helpful for large-scale labeled datasets.

\begin{table*}[h!]
\begin{footnotesize}
\bgroup
\def\arraystretch{1.75}
\begin{tabular}{p{0.2\linewidth}|ccccc}
Dataset name           & \# people        & \# images & \# pic./person & ethnic diversity              & acquisition                           \\ \hline
MS-Celeb-1M            & 100K             & \textbf{8.2M}      & 82               & uneven                        & mostly American and British actors    \\[-0.19cm]
VGGFace2               & 9.1K             & 3.3M      & 362              & uneven                        & Google Image search                   \\[-0.19cm]
CASIA-WebFace          & 10K              & 494K      & 49               & uneven                        & celebrities from IMDb                 \\[-0.19cm]
CASIA-Face-Africa      & 1.1K             & 38K       & 34               & all African                   & controlled, indoor and outdoor        \\[-0.19cm]
MegaFace (unavailable) & 672K             & 4.7M      & 7                & uneven                        & Yahoo Flickr website search queries             \\[-0.19cm]
DiveFace               & 24K              & 72K       & 3                & balanced across 3 ethnicities & subset of MegaFace                    \\[-0.19cm]
BUPT-BalancedFace      & 28K              & 1.3M      & 46               & balanced across 4 ethnicities & celebrities from MS-Celeb-1M          \\[-0.19cm]
BUPT-GlobalFace        & 38K              & 2M        & 52               & matches global population     & celebrities from MS-Celeb-1M          \\[-0.19cm]
BUPT-TransferFace      & \textgreater 10K & 600K      & 60               & 75\% Cauc. vs. others         & celebrities from MS-Celeb-1M          \\ \hline
AfricanFaceSet-5M      & \textbf{5M}               & 5M        & unlabeled    & African majority           & random faces from YouTube news videos \\[-0.21cm]
AsianFaceSet-3M        & \textbf{3M}               & 3M        & unlabeled    & Asian majority             & random faces from YouTube news videos
\end{tabular}\newline
\egroup
\end{footnotesize}
\vspace{-0.3cm}
\caption{Overview of the publicly available training sets in facial domain. Typically, the quality of a face recognition dataset is described by several factors: the number of people in the dataset, the number of images per person, and the diversity of capture conditions. While our \textit{AfricanFaceSet-5M} and \textit{AsianFaceSet-3M} datasets are not  directly suitable for face recognition due to the absence of identity labels,  they comprise more distinct people than existing large-scale datasets and contain a more diverse distribution of faces than only celebrities.}
\label{table:train-datasets}
\vspace{-0.5cm}
\end{table*}

 \vspace{-0.2cm}
\section{Related work}
\label{sec:related}

\textbf{Face recognition datasets.}
Several large-scale datasets of faces with identity labels have been released publicly, 
such as CASIA-WebFace \cite{Yi2014learning}, VGGFace2 \cite{Cao2018Vggface2}, MS-Celeb-1M \cite{Guo2016celeb} and the very recent million-scale WebFace-42M dataset~\cite{Zhu2021webface}.
However, these datasets have been collected ``in the wild'', hence, they inevitably suffer
from biases in terms of age, gender, or race.  
To better understand and stimulate research in fairness about face recognition, several datasets have been proposed. Examples of such labeled datasets include BUPT-Globalface~\cite{Wang2020mitigating} (2M images with ethnic distribution matching the world population) and BUPT-Balancedface~\cite{Wang2020mitigating} (1.3M images with the perfect ethnic split). Racial Faces in-the-Wild (RFW)~\cite{Wang2019racial} is a verification database that has been constructed from MS-Celeb-1M. Currently, it serves a standard fairness benchmark. Another \linebreak verification dataset BFW has been introduced in \cite{Robinson2020face}, which, similarly to RFW, contains eight subgroups balanced across gender and ethnicity. 
The main feature of all these datasets -- predefined ethnic split -- allows one to disambiguate dataset bias and model bias for more precise methods evaluation.
Still, the currently available data possesses a number of limitations. For instance, the number of distinct people is typically limited and is often many orders of magnitude smaller than the dataset size. Additionally, both identification and verification datasets require many images per person, which usually restricts the construction of open-source datasets to the search of celebrity pictures by text queries. Our two photo collections, released to the public, -- \textit{AfricanFaceSet-5M} and \textit{AsianFaceSet-3M}, -- fill a different gap in the space of available datasets. On the one hand, they neither have identity labels nor  feature many images per person. On the other hand, these collections are large and focused on groups and conditions that are underrepresented in general face recognition datasets. Our evaluation dataset \textit{RB-WebFace}, assembled from large-scale WebFace-42M, is a new verification dataset containing a significantly larger number of pairs and does not use external models to select the negatives (which introduces certain selection bias) compared to RFW, which is currently used as the main benchmark in the bias mitigation-focused branch of works.

\begin{figure*}[t]
    \vspace{-0.3cm}
    \begin{center}
        \centering
\includegraphics[trim=2.0cm 8.2cm 3.8cm 1.5cm,clip,width=1.0\textwidth]{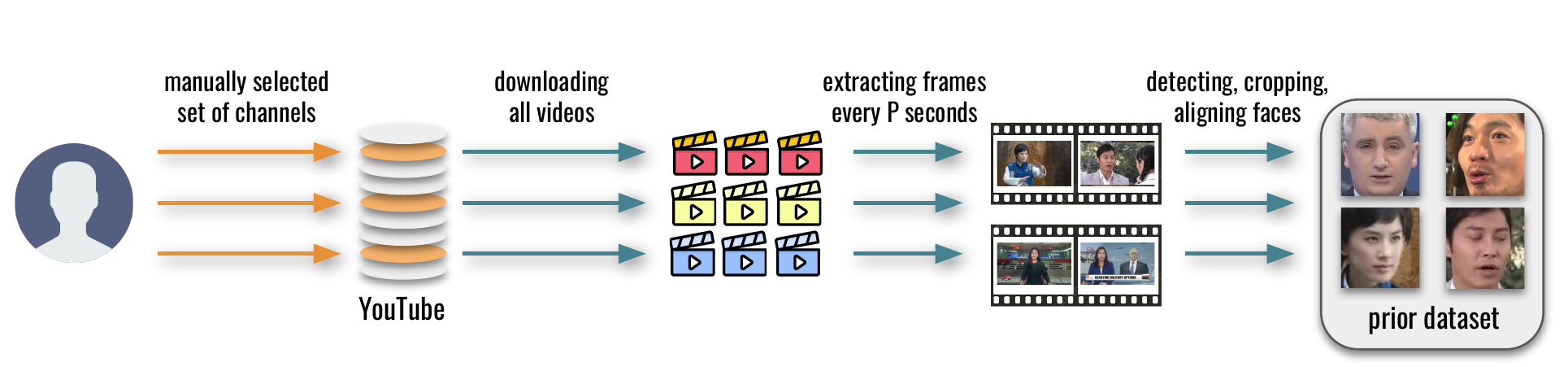}
    \end{center}\vspace{-0.5cm}
    \caption{Our data collection process starts by manually specifying a set of YouTube channels with a specific topic, e.g. a set of  news channels of a desired town or part of the world. All videos are downloaded in the highest available quality, one frame per $P$ seconds is extracted from each video, and all the faces found in the extracted frames are cropped, aligned by landmarks and resized to the target resolution. This way, we obtain millions of random faces with a desired demographic distribution.}
    \label{fig:scraping_scheme}
    \vspace{-0.4cm}
\end{figure*}

\noindent \textbf{Data augmentation}. Generating synthetic data is a possible solution to improving the performance of face recognition and in some cases reducing the negative effect of the dataset bias. An idea of this kind is pursued in \cite{marriott20213d}, a non-linear 3DMM texture model is proposed to produce sharp renderings of faces from novel poses. This technique improves generalization with respect to the head pose and illumination. A number of approaches synthesize data via face generative models. For instance, in SynFace~\cite{qiu2021synface} random face images are constructed by a GAN with identity control, and labels are constructed by a procedure similar to MixUp augmentation~\cite{zhang2017mixup}.
Similarly, in \cite{Georgopoulos2021mitigating} a  
style transfer GAN is used to simultaneously transfer multiple facial demographic attributes and generate diverse images for each attribute class. Authors of Virface~\cite{li2021virface} suggested a method for incorporating additional negative (impostor) pairs from the unlabeled data showing a boost in metrics. 
Zhang et al. \cite{zhang2021applicability} study the applicability of the data generated by StyleGAN by inspecting the distributions of downstream models but do not study its effect on face recognition metrics. 

Our approach is inspired by the latter idea of generating synthetic samples but takes a step forward by utilizing the generative model itself (StyleGAN with an encoder in our case). As shown in~\cite{Richardson2021encoding,Tov2021designing,Alaluf2021restyle,Alaluf2021hyperstyle}, the StyleGAN encoders have high potential for both generating new realistic faces from a latent code and solving the inversion problem. This highlights the expressiveness of these models and richness of their internal representations.

\noindent \textbf{Self-supervised pretraining.} Despite the general dominance of supervised learning in practical ML and CV, approaches based on self-supervised learning have been evolving in various forms, such as self-organizing or siamese networks~\cite{Becker1992self,Bromley1993signature,Chopra2005learning,Hadsell2006dimensionality}.
Currently, self-supervised learning is the dominant approach in NLP with a wide spectrum of possible approaches~\cite{Liu2019roberta,Conneau2019unsupervised,Clark2020electra} and is being actively adapted in computer vision~\cite{Dosovitskiy2020image}. 

Analogously, we are witnessing the first attempts to apply self-supervised learning now being integrated into the face recognition frameworks. In 3D-BERL~\cite{he2022enhancing}, the performance over multiple benchmarks is improved by a separate 3D reconstruction network branch. The work~\cite{lin2021domain} studies the effect of self-supervised learning for domain transfer in face recognition. Incremental learning~\cite{wu2019large} can be bridged with self-supervised approaches to adapt to large number of target classes~\cite{zhang2020self}. The procedure for collecting large-scale unlabeled datasets allows us to adapt self-supervised training in a more conventional fashion while proving its efficacy for our application.

 \vspace{-0.1cm}
\section{Method}

Our pipeline is comprised of several stages. First, we train StyleGAN2-ADA to fit the face distribution on the unlabeled dataset. Second, the pSp encoder is trained (also on the unlabeled dataset), that will define a feature extractor well-suited for the group of interest. Finally, the encoder is fine-tuned for the downstream face recognition task. The entire procedure is outlined in Subsec.~\ref{subsec:architecture} and visually described in Fig.~\ref{fig:method_scheme}. Before describing the method itself, we outline the schematics of our prior dataset collection procedures in Subsec.~\ref{subsec:prior_dataset}.

\subsection{Prior dataset collection}
\label{subsec:prior_dataset}

Due to the nature of the task, the prior dataset must contain the samples from the group of interest. At the same time, samples from the prior dataset do not require identity labels of any kind, unlike samples of the face recognition datasets. In practice, this removes the restriction of linking photos of the same person (something that might be considered a violation of privacy), and thus enlarges the search space and simplifies data collection. Still, fulfilling several requirements for the prior dataset remains a challenge, such as: collecting faces only from a specific group (e.g. an ethnic or gender group), obtaining a large number of them (preferably, an order of magnitude more than in the face recognition dataset to perform subsequent transfer learning), and using only data legally allowed for collection. We found semi-automatic YouTube channel scraping to be an efficient solution that satisfies these requirements. In particular, we propose to select a set of publicly available YouTube channels dedicated to the desired group of interest. The channel names are the only entry point and the only manually performed step for the scraping procedure (see Fig.~\ref{fig:scraping_scheme}). Typically, the requirement for the channels is having them systematically featuring different people; a possible example would be a set of news channels released in a country of choice. All videos are downloaded from every channel, one frame per $P$ seconds of each video is extracted, and all faces are detected, cropped and aligned by landmarks via MTCNN~\cite{Zhang2016joint} library. Our data has been scraped from a predefined set of channels such as news channels and others.

The unlabeled prior dataset will later be referred to as $\mathcal{D}^{prior}$. Table~\ref{table:train-datasets} describes the relative difference of the collected data to the datasets typically used for face recognition. Note however, that despite the latter being directly inapplicable for training due to the lack of the labels, they potentially contain more people than the others.

\subsection{Architecture and the training procedure}
\label{subsec:architecture}

The central part of our pipeline is a single face recognition convnet $f_{\theta, \psi}(I)$ (a backbone), that takes a single RGB image $I$ as an input, and outputs a 512-dimensional vector $e \in \mathbb{R}^{512}$ (face embedding). Typically, the backbone is trained on a training dataset with angular margin based losses, such as SphereFace~\cite{Liu2017sphereface}, ArcFace~\cite{Deng2019arcface}, and others, in standard works on face recognition. In our pipeline, we follow the same procedure, but only after a pretraining stage is performed. In order to do so, we first train the StyleGAN2-ADA~\cite{Karras2020training} generator $g_\phi(w(z))$ that transforms a latent vector $z \in \mathcal{Z} \subseteq \mathbb{R}^{512}$ into \textit{unfolded} latent space $w(z) \in \mathcal{W}+ \subseteq \mathbb{R}^{L \times 512}$, and then into a realistic face image $\hat{I}$. The generator $g_\phi(w(z))$ is trained together with a discriminator that allows the generator to learn the distribution of faces in $\mathcal{D}^{prior}$. The training procedure for StyleGAN2-ADA follows the one in the corresponding paper. Secondly, we introduce a network $f'_{\theta, \omega}(x)$, which is trained as an encoder for the StyleGAN generator, i.e. a network that solves an inverse problem: given an RGB image $I$, predict an unfolded latent code $w \in \mathcal{W}+$, such that the corresponding reconstruction  $\hat{I} = g_\phi(w)$ is as close to the input $I$ as possible. A set of different possible approaches has appeared recently for training an encoder for StyleGAN2, ranging from applying a ConvNet with style-predicting layers~\cite{Richardson2021encoding} to employing hyper-networks~\cite{Alaluf2021restyle}. In our method, the encoder architecture follows the pSp method~\cite{Richardson2021encoding}, that proposes predicting a latent code by a ConvNet divided into a convolution part (with parameters $\theta$), following a ResNet architecture, and a set of fully-convolutional style predictors -- \textit{map2style} blocks (with parameters $\omega$). The training procedure for the encoder follows the one outlined in~\cite{Richardson2021encoding} with a few variations. Namely, we only keep the fidelity losses (standard L2 and neural-based LPIPS~\cite{Zhang2018unreasonable}) and disable the identity loss proposed by the authors.
During the encoder training, the StyleGAN generator remains frozen in order to fix the latent space. A pair of networks $f'_{\theta, \omega}(x)$ and $g_\phi(w(z))$ can be seen as an asymmetric autoencoder (since the number of encoder parameters highly surpasses the number of generator parameters) with the decoder pretrained for a generative task and the encoder subsequently trained for a discriminative (regressive) task. The visual quality and the identity preservation of the reconstruction $\hat{I}$ indirectly defines the expressiveness of the latent code $l$ predicted by the encoder.

Finally, after the encoder $f'_{\theta, \omega}(I)$ is trained, we transfer its convolutional parameters $\theta$ into the face recognition backbone $f_{\theta, \psi}(I)$, which also comprises a set of new parameters $\psi$, corresponding to a fully-connected layer in the end of the network. We repeat the standard training procedure for face recognition, e.g. described in ArcFace~\cite{Deng2019arcface} and others. The backbone is trained on a face recognition dataset $\mathcal{D}^{facerec} = \{ \mathcal{D}_1, \dots, \mathcal{D}_N \}$, where each group $\mathcal{D}_i$ corresponds to a set of images of the same person \#$i$. 
The slight differences in hyperparameters that we found beneficial when used with pretraining and technical details are described in the Appendix.

 \section{Results}
\label{sec:results}

\subsection{Data and evaluation protocol}
\label{subsec:impl}

\begin{table}[h]
\vspace{-0.5cm}
\begin{center}
\resizebox{\linewidth}{!}{
\begin{tabular}{l|cccc|cc}
                                                                  & \multicolumn{4}{c|}{RFW, accuracy \%, ↑}                          & \multicolumn{1}{l}{} & \multicolumn{1}{l}{} \\ \hline
                                                                  & Cauc.      & African        & Asian          & Indian         & Avg ↑                & Std ↓                \\ \hline
ArcFace R-50*~\cite{Deng2019arcface}                                                     & 96.18          & 93.98          & 93.72          & 94.67          & 94.64                & 1.11                 \\
CosFace R-50*~\cite{Wang2018cosface}                                                     & 95.12          & 93.93          & 92.98          & 92.93          & 93.74                & 0.89                 \\
DebFace*~\cite{Gong2019debface}                                                          & 95.95          & 93.67          & 94.33          & 94.78          & 94.68                & 0.83                 \\
ACNN*~\cite{Kang2017incorporating}          & 96.12          & 94.00          & 93.67          & 94.55          & 94.58                & 0.94                 \\
PFE*~\cite{Shi2019probabilistic}            & 96.38          & \textbf{95.17} & 94.27          & 94.60          & 95.11                & 0.93                 \\
RL-RBN*~\cite{Wang2020mitigating}           & 96.27          & 95.00          & \textbf{94.82}          & 94.68          & \textbf{95.19}                & \textbf{0.63}                 \\
GAC R-50*~\cite{Gong2021mitigating} & 96.27          & 94.40          & 94.32 & 94.77 & 94.94                & 0.79        \\ \hline \hline
Baseline (ArcFace)                                                      & 96.00          & 94.00          & 93.08          & 94.48          & 94.39                & 1.06                 \\
\textcolor{pretrainingclr}{+ $\mathcal{D}^{prior}$ (African)}                  & 96.35          & 94.37          & 93.62          & 94.88          & 94.81                & 1.00                 \\
\textcolor{pretrainingclr}{+ $\mathcal{D}^{prior}$ (Asian)}                    & 96.38          & 94.67          & \underline{94.03}          & \underline{\textbf{95.03}}          & 95.03                & \underline{0.86}                 \\
\textcolor{pretrainingclr}{+ $\mathcal{D}^{prior}$ (Afr+Asian)}                      & \underline{\textbf{96.52}} & \underline{95.00}          & 93.90          & 94.93          & \underline{95.09}                & 0.94                 \\ \hline
\multicolumn{1}{|r|}{\textcolor{gapclr}{increase}}                      & \multicolumn{1}{r}{+0.52} & \multicolumn{1}{r}{\textcolor{pretrainingclr}{+1.00}}          & \multicolumn{1}{r}{\textcolor{pretrainingclr}{+0.82}}          & \multicolumn{1}{r|}{+0.45}          &                 &     \multicolumn{1}{r|}{}             \\ \hline
\end{tabular}
}\newline
\caption{Comparison of the verification accuracy of the methods on RFW validation set. Asterisk ``*'' indicates that the numbers are directly taken from the respective table in~\cite{Gong2021mitigating}. In case of the ArcFace R-50 baseline, it also corresponds to a slightly different reported training procedure. For other methods, the epoch with the best RFW African accuracy score is taken. For GAC~\cite{Gong2021mitigating}, we consider the experiment with \textit{Estimated} ethnic labels in order to compare in the same setting when ground truth labels are not given as an input to the model. The \textcolor{pretrainingclr}{pretraining scheme} is efficient when applied to the common choice of the FR method (ArcFace) and the best quality \textcolor{gapclr}{increase} is for the ethnicities seen during pretraining (African and Asian).} \label{table:main}
\end{center}
\vspace{-0.7cm}
\end{table}

\begin{table}[h!]
\resizebox{\linewidth}{!}{
\begin{tabular}{l|cccc|}
                                & \multicolumn{4}{c|}{RB-WebFace}  \\
                                & \multicolumn{4}{c|}{TPR @ FPR=$10^{-3}$ ↑}   \\ & Cauc.         & African           & \multicolumn{1}{c}{Asian} & \multicolumn{1}{c|}{Indian} \\ \hline
ArcFace                    &    89.86    &    86.73    &    94.31    &    \multicolumn{1}{c|}{93.82}     \\
ArcFace \textcolor{pretrainingclr}{+ $\mathcal{D}^{prior}$ (African)}                             & 92.62    &    90.20    &    96.31    &    \multicolumn{1}{c|}{95.85}          \\
ArcFace \textcolor{pretrainingclr}{+ $\mathcal{D}^{prior}$ (Asian)}                           & 92.91    &    90.61    &    96.15    &    \multicolumn{1}{c|}{95.83}    \\
ArcFace \textcolor{pretrainingclr}{+ $\mathcal{D}^{prior}$ (Afr+Asian)}  & \underline{93.54}    &    \textbf{\underline{91.51}}    &    \underline{96.51}    &    \multicolumn{1}{c|}{\textbf{\underline{96.17}}}    
    \\ \hline
\multicolumn{1}{|r|}{\textcolor{gapclr}{increase}}                      & \multicolumn{1}{r}{+3.68} & \multicolumn{1}{r}{\textcolor{pretrainingclr}{+4.78}}          & \multicolumn{1}{r}{\textcolor{pretrainingclr}{+2.20}}          & \multicolumn{1}{r|}{+2.35}    \\ \hline\hline
SphereFace               &  92.08  & 89.51  &  95.52  & \multicolumn{1}{c|}{95.02}                       \\ 
SphereFace \textcolor{pretrainingclr}{+ $\mathcal{D}^{prior}$ (Afr+Asian)}              &  \underline{92.83}  &  \underline{90.52} & \underline{96.22} & \multicolumn{1}{c|}{\underline{95.65}}   \\ \hline   
\multicolumn{1}{|r|}{\textcolor{gapclr}{increase}}                      & \multicolumn{1}{r}{+0.75} & \multicolumn{1}{r}{\textcolor{pretrainingclr}{+1.01}}          & \multicolumn{1}{r}{\textcolor{pretrainingclr}{+0.70}}          & \multicolumn{1}{r|}{+0.63}    \\ \hline\hline
GAC               &  92.48    &    89.13  & 96.33    & \multicolumn{1}{c|}{94.66}                       \\ 
GAC \textcolor{pretrainingclr}{+ $\mathcal{D}^{prior}$ (Afr+Asian)}               & \textbf{\underline{93.60}}    &    \underline{90.80}  & \textbf{\underline{96.76}}    & \multicolumn{1}{c|}{\underline{95.46}}                      \\ \hline
\multicolumn{1}{|r|}{\textcolor{gapclr}{increase}}                      & \multicolumn{1}{r}{+1.12} & \multicolumn{1}{r}{\textcolor{pretrainingclr}{+1.67}}          & \multicolumn{1}{r}{\textcolor{pretrainingclr}{+0.43}}          & \multicolumn{1}{r|}{+0.80}    \\ \hline\hline
\end{tabular}
}\newline\\[0.1cm]
\resizebox{\linewidth}{!}{
\begin{tabular}{l|cccc|}
                                & \multicolumn{4}{c|}{TPR @ FPR=$10^{-4}$ ↑}   \\  \hline
ArcFace                   &    81.48    &    76.80    &    87.47    &    86.89   \\
ArcFace \textcolor{pretrainingclr}{+ $\mathcal{D}^{prior}$ (African)}                        &    85.56    &    81.99    &    91.42    &    90.67           \\
ArcFace \textcolor{pretrainingclr}{+ $\mathcal{D}^{prior}$ (Asian)}                          &    85.79    &    82.41    &    91.12    &    90.36        \\
ArcFace \textcolor{pretrainingclr}{+ $\mathcal{D}^{prior}$ (Afr+Asian)}   &    \underline{86.84}    &    \textbf{\underline{83.74}}    &    \underline{91.74}    &    \textbf{\underline{91.19}}
    \\ \hline
\multicolumn{1}{|r|}{\textcolor{gapclr}{increase}}                      & \multicolumn{1}{r}{+5.36} & \multicolumn{1}{r}{\textcolor{pretrainingclr}{+6.94}}          & \multicolumn{1}{r}{\textcolor{pretrainingclr}{+4.27}}          & \multicolumn{1}{r|}{+4.30}    \\ \hline\hline
SphereFace               &  83.63  & 80.04  & 89.10   & \multicolumn{1}{c|}{89.07}                       \\ 
SphereFace \textcolor{pretrainingclr}{+ $\mathcal{D}^{prior}$ (Afr+Asian)}              &  \underline{85.14}  & \underline{81.47}  &  \underline{90.65}  & \multicolumn{1}{c|}{\underline{90.16}}
                 \\ \hline
\multicolumn{1}{|r|}{\textcolor{gapclr}{increase}}                      & \multicolumn{1}{r}{+1.51} & \multicolumn{1}{r}{\textcolor{pretrainingclr}{+1.43}}          & \multicolumn{1}{r}{\textcolor{pretrainingclr}{+1.55}}          & \multicolumn{1}{r|}{+1.09}    \\ \hline\hline
GAC                             &      85.64    &     80.24      &    91.72           &   88.12     \\ 
GAC \textcolor{pretrainingclr}{+ $\mathcal{D}^{prior}$ (Afr+Asian)}             &      \textbf{\underline{86.91}}    &     \underline{82.47}      &    \textbf{\underline{92.39}}           &   \underline{89.48}     \\ \hline
\multicolumn{1}{|r|}{\textcolor{gapclr}{increase}}                      & \multicolumn{1}{r}{+1.27} & \multicolumn{1}{r}{\textcolor{pretrainingclr}{+2.23}}          & \multicolumn{1}{r}{\textcolor{pretrainingclr}{+0.67}}          & \multicolumn{1}{r|}{+1.36}    \\ \hline\hline
\end{tabular}
}\newline
\caption{Comparison of the methods on
the newly assembled RB-WebFace validation set. Publicly available authors' implementation of GAC~\cite{Gong2021mitigating} has been used to retrain the method and evaluate the quality. Here, we showcase the TPR given two pre-selected FPR thresholds to highlight the error rate difference separately for each ethnic group. For GAC, the setting \textit{GAC (Estimated)} (see~\cite{Gong2021mitigating}) is used. This table indicates how the proposed \textcolor{pretrainingclr}{pretraining scheme} provides an additional boost for different pipelines.}
\label{table:webface}
\vspace{-0.3cm}
\end{table}

\begin{table}[h!]
\centering
\resizebox{.8\linewidth}{!}{
\begin{tabular}{l|cccc|}
                                & \multicolumn{4}{c|}{RB-WebFace}                                                                                                                                                                      \\
                                & \multicolumn{4}{c|}{TPR @ FPR=$10^{-3}$ ↑}                  \\ & Cauc.         & African           & Asian             & \multicolumn{1}{c|}{Indian}           \\ \hline
Baseline                    &    89.86    & 86.73 & 94.31 & \multicolumn{1}{c|}{93.82}                    \\
\textcolor{pretrainingclr}{+ 1\% $\mathcal{D}^{prior}$}  & 92.59 & 90.44 & 96.06 & \multicolumn{1}{c|}{95.72}        \\ 
\textcolor{pretrainingclr}{+ 10\% $\mathcal{D}^{prior}$}  & 92.84 & 90.72 & 96.12 & \multicolumn{1}{c|}{95.64}      \\
\textcolor{pretrainingclr}{+ 100\% $\mathcal{D}^{prior}$}  & \underline{\textbf{93.54}} & \underline{\textbf{91.51}} & \underline{\textbf{96.51}} & \multicolumn{1}{c|}{\underline{\textbf{96.17}}}       \\  \hline
\end{tabular}
}
\caption{Ablation study of the quality dependence on the $|\mathcal{D}^{prior}|$ prior dataset size on the newly assembled RB-WebFace validation set. $\mathcal{D}^{prior}$ = \{AfricanFaceSet-5M $\cup$ AsianFaceSet-3M\} was used in this set of experiments.}
\label{table:rb_webface_wrt_prior_dset_size}
\vspace{-0.5cm}
\end{table}

Our system requires two datasets -- a labeled face recognition dataset $\mathcal{D}^{facerec}$ and an unlabeled prior dataset $
\mathcal{D}^{prior}$. As for the $\mathcal{D}^{facerec}$, we employ BUPT-BalancedFace~\cite{Wang2020mitigating} due to its ethnic and gender balance guarantees. Namely, it consists of 1.3 million images belonging to 28K different people, divided into 4 ethnic groups of equal size -- 7K African, East Asian, Indian, and Caucasian people each. Other state-of-the-art methods can only take advantage of a face recognition dataset, 
and compare with them on BUPT-BalancedFace.

The prior dataset $\mathcal{D}^{prior}$ for our work has been collected from YouTube via the procedure outlined in Subsec.~\ref{subsec:prior_dataset}. The dataset consists of two parts -- African and East Asian ethnic groups -- corresponding to two of the four benchmarks, these groups demonstrate higher error rates for face recognition systems compared to others in the branch of works~\cite{Gong2021mitigating,Wang2020mitigating,Gong2019debface}. Analogously, these two groups are known to be challenging according to NIST evaluation on demographics~\cite{Grother2019Face} and benchmarks of other tasks~\cite{Buolamwini2018gender}. As an input for the scraping routine, we have used a set of 30-40 news channels corresponding to the respective part of the world. One frame per $P=5$ seconds was taken, and only frames from the start through the 20th minute of each video have been considered. Since the target resolution for our network training is $112 \times 112$, we only approve face bounding boxes selected by MTCNN if the face occupies at least 100 px on each side. Additionally, we increase the decision thresholds for MTCNN to reduce the number of false positives during the detection process.  The exact set of the news channels is enlisted in the Appendix. In total, 5 million images for the African group (\textit{AfricanFaceSet-5M}) and 3 million images for the East Asian group (\textit{AsianFaceSet-3M}) have been collected. 

One face recognition validation protocol is based on a balanced verification set Racial Faces In-the-Wild (RFW)~\cite{Wang2019racial} (following ~\cite{Xu2021consistent,Gong2021mitigating,Gong2019debface,Wang2020mitigating}). It consists of 3K positive pairs of samples (pairs of images of the same person) and 3K negative pairs (pairs of images of different but similarly looking people, selected by a face recognition algorithm which can introduce selection bias) for each ethnic group, summing up to 24K pairs. Evaluation is carried out in an LFW-like protocol~\cite{Huang2008labeled} that involves evaluating the backbone $f_{\theta, \psi}(I)$ on images of each pair and thresholding the resulting cosine distances between embeddings. Each set of 3K+3K pairs is constructed from 3K people comprising a subset of MS-Celeb-1M~\cite{Guo2016celeb}. 

In addition to that, we propose a new test set \textit{RB-WebFace} constructed in a similar fashion from the recently proposed million-scale identification dataset \textit{WebFace-42M} (a cleaned version of \textit{WebFace-260M}~\cite{Zhu2021webface}). \textit{RB-WebFace} is constructed by evaluating a pretrained ethnic group classifier on WebFace-42M to separate it into four ethnic groups (the group assignment is later refined by a consensus procedure). For each group, the largest possible number of distinct people is taken to construct positive and negative pairs. 
The details are provided in Appendix, as well as the example pictures and comparison to typically used test datasets by the number of people and pairs.

\begin{figure*}[t!]
     \centering
     \begin{subfigure}[b]{0.32\linewidth}
         \centering
         \includegraphics[trim={0.6cm 0.7cm 0.4cm 0.5cm},clip,width=\textwidth]{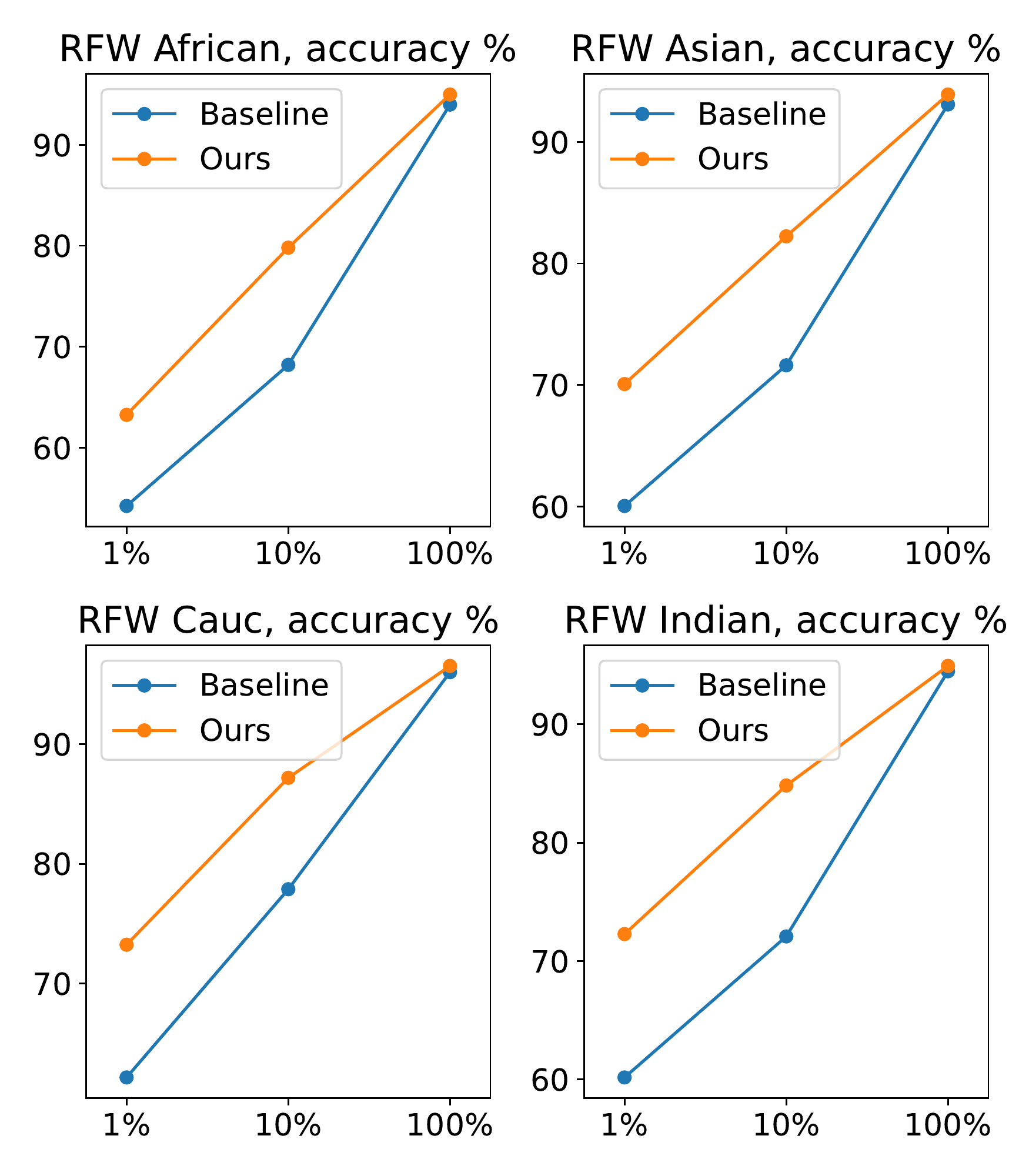}
     \end{subfigure}
     \hfill
     \begin{subfigure}[b]{0.32\linewidth}
         \centering
         \includegraphics[trim={0.6cm 0.7cm 0.4cm 0.5cm},clip,width=\textwidth]{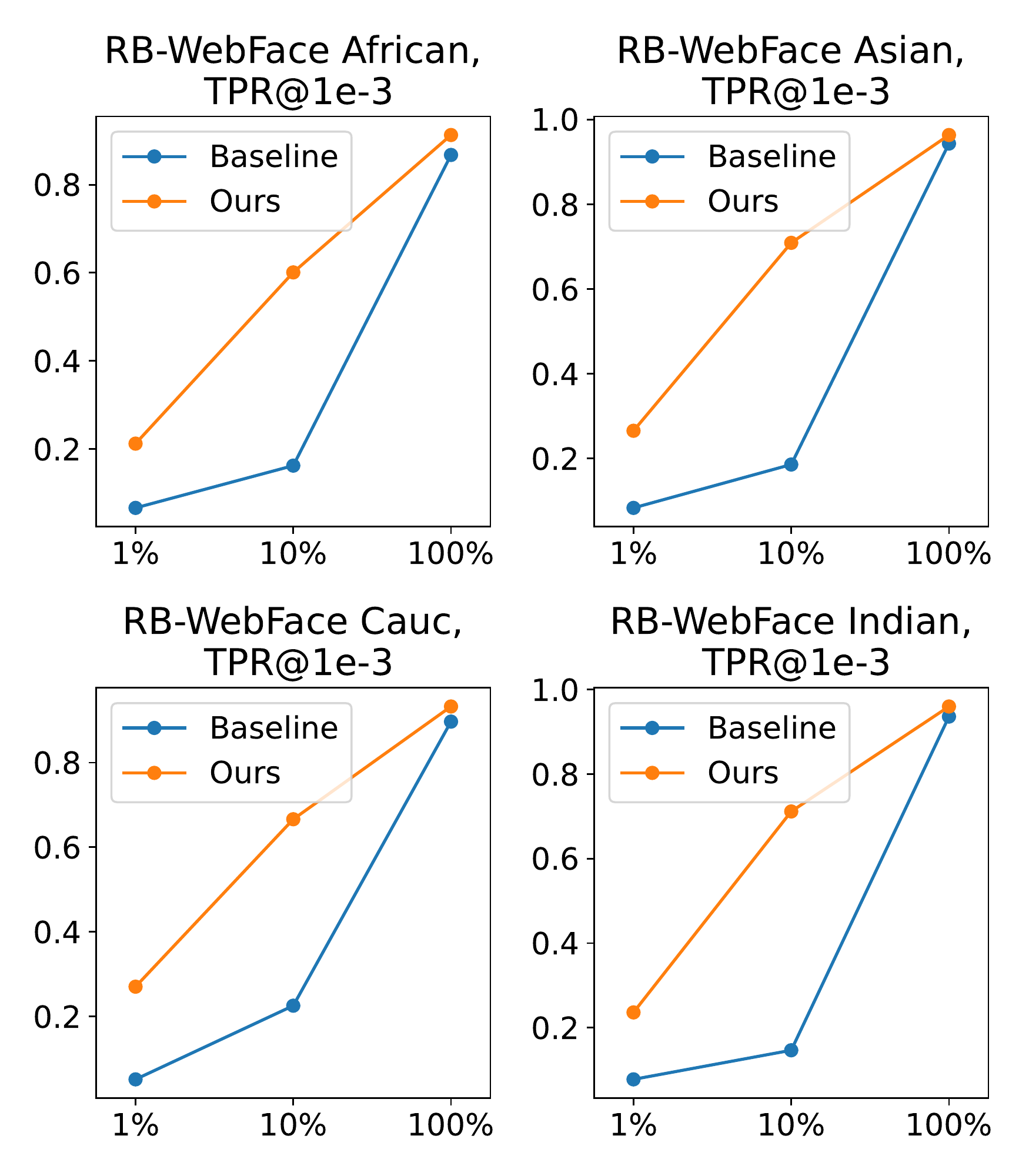}
     \end{subfigure}
     \hfill
     \begin{subfigure}[b]{0.32\linewidth}
         \centering
         \includegraphics[trim={0.6cm 0.7cm 0.4cm 0.5cm},clip,width=\textwidth]{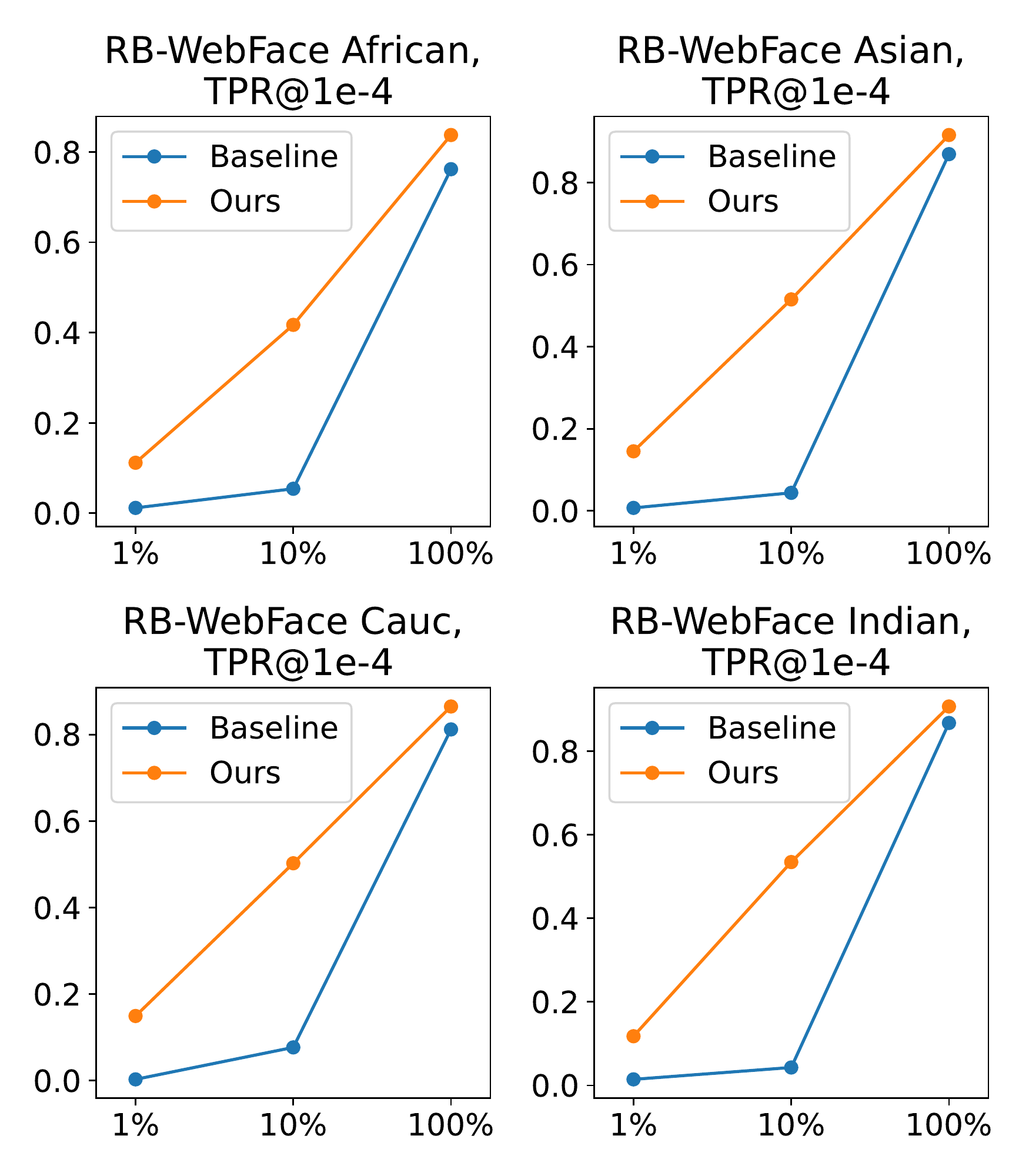}
     \end{subfigure}
    \caption{Ablation study of the dependence of the test quality, evaluated for RFW and RB-WebFace datasets, on the number of labeled samples from $|\mathcal{D}^{facerec}|$ used for fine-tuning. In this experiment, $\mathcal{D}^{prior} = \{\textrm{AfricanFaceSet-5M}\, \cup\, \textrm{AsianFaceSet-3M}\}$, while $\mathcal{D}^{facerec} = \textrm{BUPT-BalancedFace}$. For $\mathcal{D}^{facerec}$, fraction of the data (100\%, 10\%, 1\%) defines the number of uniformly sampled people left in the dataset. In the plot legend, \textit{Baseline} is ArcFace R-50.}
    \label{table:ablation_dset_size}
    \vspace{-0.3cm}
\end{figure*}

\subsection{Evaluation}

\noindent \textbf{Comparison with the state-of-the-art on RFW.} The Table~\ref{table:main} contains the comparison to baseline (ArcFace R-50) and a few state-of-the-art methods (GAC~\cite{Gong2021mitigating}, DebFace~\cite{Gong2019debface}, and others). RFW verification accuracy is reported for each of the four ethnic groups. We demonstrate an increase in accuracy for all races compared to the baseline, which is on par with the state-of-the-art and outperforming on Caucasian and Indian ethnic groups. The largest increase w.r.t. the baseline is observed for the African group. It is important to note that our pretraining scheme can also be used to initialize other methods and enhance their results in a similar fashion.

\noindent \textbf{Comparison on newly assembled RB-WebFace test set.} Since the number of positive and negative pairs is orders of magnitude different, we report the ROC curves values (TPR vs. FPR), better suited for the class-imbalanced evaluations, instead of reporting accuracy as for RFW. By sweeping the threshold (in [0.1, 0.75] range), we obtain the (TPR, FPR) pairs, and TPR @ FPR = \{$10^{-3}$, $10^{-4}$\} is reported in Table~\ref{table:webface}. Our method (denoted as \textit{Baseline + $\mathcal{D}^{prior}$}) is compared to the \textit{Baseline} (ArcFace R-50). Additionally, we report the enhancement that our procedures provides to other methods~\cite{Liu2017sphereface,Gong2021mitigating}. The main conceptual result that we demonstrate is the significant increase of the TPR vs. given FPR for all races and decrease of ethnic bias on RB-WebFace. A separate plot in the Appendix comprises the TPR measurements for a wide range of possible FPR values. We also provide the results for ResNet-\{34,100\} backbones in Table~\ref{table:rb_webface_backbone_capacity}.

\noindent \textbf{Limited labeled data.} The study in Fig.~\ref{table:ablation_dset_size} describes the dependency of our method's accuracy on the number of samples in $\mathcal{D}^{facerec}$. We demonstrate the difference between our method and the ArcFace R-50 baseline for each of the benchmarks, both on RFW and RB-WebFace. As expected, pretraining helps the most when the network is fine-tuned on more limited amounts of labeled data. This especially highlights the benefits of using self-supervised learning in these scenarios. 

\noindent \textbf{Limited prior dataset.} An ablation in Table~\ref{table:rb_webface_wrt_prior_dset_size} describes the quality dependence on the size of the prior dataset. As shown, gradually enlarging it results in a monotonous performance increase for all ethnic groups. This indicates that the use of large amounts of diverse unlabeled data significantly strengthens the method in the end.

\begin{table}[h!]
   \vspace{-0.2cm}
   \centering
    \resizebox{.85\linewidth}{!}{
       \begin{tabular}{lcccc}
        \multicolumn{1}{l|}{}         & \multicolumn{4}{c|}{RB-WebFace}                                                                  \\
        \multicolumn{1}{c|}{}         & \multicolumn{4}{c|}{TPR @ FPR=$10^{-3}$ ↑}                                                       \\
        \multicolumn{1}{l|}{}         & Cauc.                & African              & Asian                & \multicolumn{1}{c|}{Indian} \\ \hline
        \multicolumn{1}{l|}{Baseline} & 89.86                & 86.73                & 94.31                & \multicolumn{1}{c|}{93.82}  \\
        \multicolumn{1}{l|}{\textcolor{vaeclr}{+ AE on $\mathcal{D}^{prior}$}}     &   92.93        &      90.83       &    95.73      & \multicolumn{1}{c|}{95.68}       \\
        \multicolumn{1}{l|}{\textcolor{vaeclr}{+ VAE on $\mathcal{D}^{prior}$}}    &   92.45     &     90.39      &    95.64        & \multicolumn{1}{c|}{95.40}       \\ \hline
        \multicolumn{1}{l|}{\textcolor{pretrainingclr}{+ ours on $\mathcal{D}^{prior}$}}   & \textbf{93.54}                & \textbf{91.51}                & \textbf{96.51}                & \multicolumn{1}{c|}{\textbf{96.17}}  \\ \hline
                                      & \multicolumn{1}{l}{} & \multicolumn{1}{l}{} & \multicolumn{1}{l}{} & \multicolumn{1}{l}{}        \\
        \multicolumn{1}{c|}{}         & \multicolumn{4}{c|}{TPR @ FPR=$10^{-4}$ ↑}                                                       \\ \hline
\multicolumn{1}{l|}{Baseline} & 81.48                & 76.80                & 87.47                & \multicolumn{1}{c|}{86.89}  \\ 
        \multicolumn{1}{l|}{\textcolor{vaeclr}{+ AE on $\mathcal{D}^{prior}$}}     &   85.85         &     83.09      &    90.46      & \multicolumn{1}{c|}{89.86}       \\
        \multicolumn{1}{l|}{\textcolor{vaeclr}{+ VAE on $\mathcal{D}^{prior}$}}    &   85.19             &     82.08        &   90.38       & \multicolumn{1}{c|}{89.25}      \\ \hline
        \multicolumn{1}{l|}{\textcolor{pretrainingclr}{+ ours on $\mathcal{D}^{prior}$}}   & \textbf{86.84}                & \textbf{83.74}                & \textbf{91.74}                & \multicolumn{1}{c|}{\textbf{91.19}}  \\ \hline
        \end{tabular}
    }
    \caption{Comparison of different pretraining strategies. Along with \textcolor{pretrainingclr}{ours} based on consequent training of \textcolor{pretrainingclr}{StyleGAN and a ResNet encoder}, one could employ \textcolor{vaeclr}{vanilla (AE)} and \textcolor{vaeclr}{variational (VAE)} autoencoders for pretraining (with the same ResNet-50 architecture as an encoder). Despite simpler simultaneous training of an encoder and a decoder in a single stage, the StyleGAN-based procedure is a more powerful prior. \textit{Baseline} refers to the ArcFace R-50.}
    \label{table:vae}
    \vspace{-0.1cm}
\end{table}

\begin{table}[h!]
\resizebox{\linewidth}{!}{
\begin{tabular}{l|cccccccc|}
                                & \multicolumn{8}{c|}{RB-WebFace} \\
                                & \multicolumn{4}{c|}{TPR @ FPR=$10^{-3}$ ↑}                                                             & \multicolumn{4}{c|}{TPR @ FPR=$10^{-4}$ ↑}    \\ & Cauc.         & Afr.           & Asian             & \multicolumn{1}{c|}{Indian}            & Cauc.         & Afr.           & \multicolumn{1}{c}{Asian} & \multicolumn{1}{c|}{Indian}  \\ \hline
R-34                   & 84.57  & 81.47 & 91.15  & \multicolumn{1}{c|}{89.91} & 73.60 & 69.23  & 83.11  & 80.40   \\
\textcolor{pretrainingclr}{+ $\mathcal{D}^{prior}$} & \textbf{90.92}    &    \textbf{88.34}    &    \textbf{95.27}    &    \multicolumn{1}{c|}{\textbf{94.39}} & \textbf{83.08}   & \textbf{78.69} &  \textbf{89.25}  & \textbf{87.98}      \\ \hline
R-50    &   89.86                & 86.73                & 94.31                & \multicolumn{1}{c|}{93.82}  & 81.48                & 76.80                & 87.47                & \multicolumn{1}{c|}{86.89} \\
\textcolor{pretrainingclr}{+ $\mathcal{D}^{prior}$}& \textbf{93.54}    &    \textbf{91.51}    &    \textbf{96.51}    &    \multicolumn{1}{c|}{\textbf{96.17}} & \textbf{86.84}   & \textbf{83.74} &  \textbf{91.74}  & \textbf{91.19}      \\ \hline
R-100                   &  93.77 & 91.55  &  96.32 &  \multicolumn{1}{c|}{96.62}  & 87.28  & 84.28  & 91.46  &  91.66   \\
\textcolor{pretrainingclr}{+ $\mathcal{D}^{prior}$}& \textbf{94.82}    &    \textbf{93.00}    &    \textbf{96.70}    &    \multicolumn{1}{c|}{\textbf{96.92}} & \textbf{88.76}   & \textbf{86.00} &  \textbf{92.24}  & \textbf{92.70}      \\ \hline
\end{tabular}
}\newline
\caption{Comparison of different ResNet backbones on RB-WebFace. The baseline is ArcFace and $\mathcal{D}^{prior}$ = AfricanFaceSet-5M $\cup$ AsianFaceSet-3M.}
\label{table:rb_webface_backbone_capacity}
\vspace{-0.5cm}
\end{table}

\noindent \textbf{Can StyleGAN be replaced with a simpler encoder-decoder architecture?} In Table~\ref{table:vae}, we compare our pretraining procedure to more conventional ones where the ResNet-50 encoder is pretrained in a symmetric autoencoder (AE) or in a symmetric variational autoencoder (VAE) setting. 
While these models can be pretrained in a single stage, our proposed StyleGAN+encoder approach requires two separate pretraining stages but demonstrates better performance, which motivates the use of state-of-the-art generative models for face recognition.

\begin{figure}[]
    \centering
    \vspace{-0.51cm}
    \includegraphics[trim={1cm 0 0 0 0},width=0.6\linewidth]{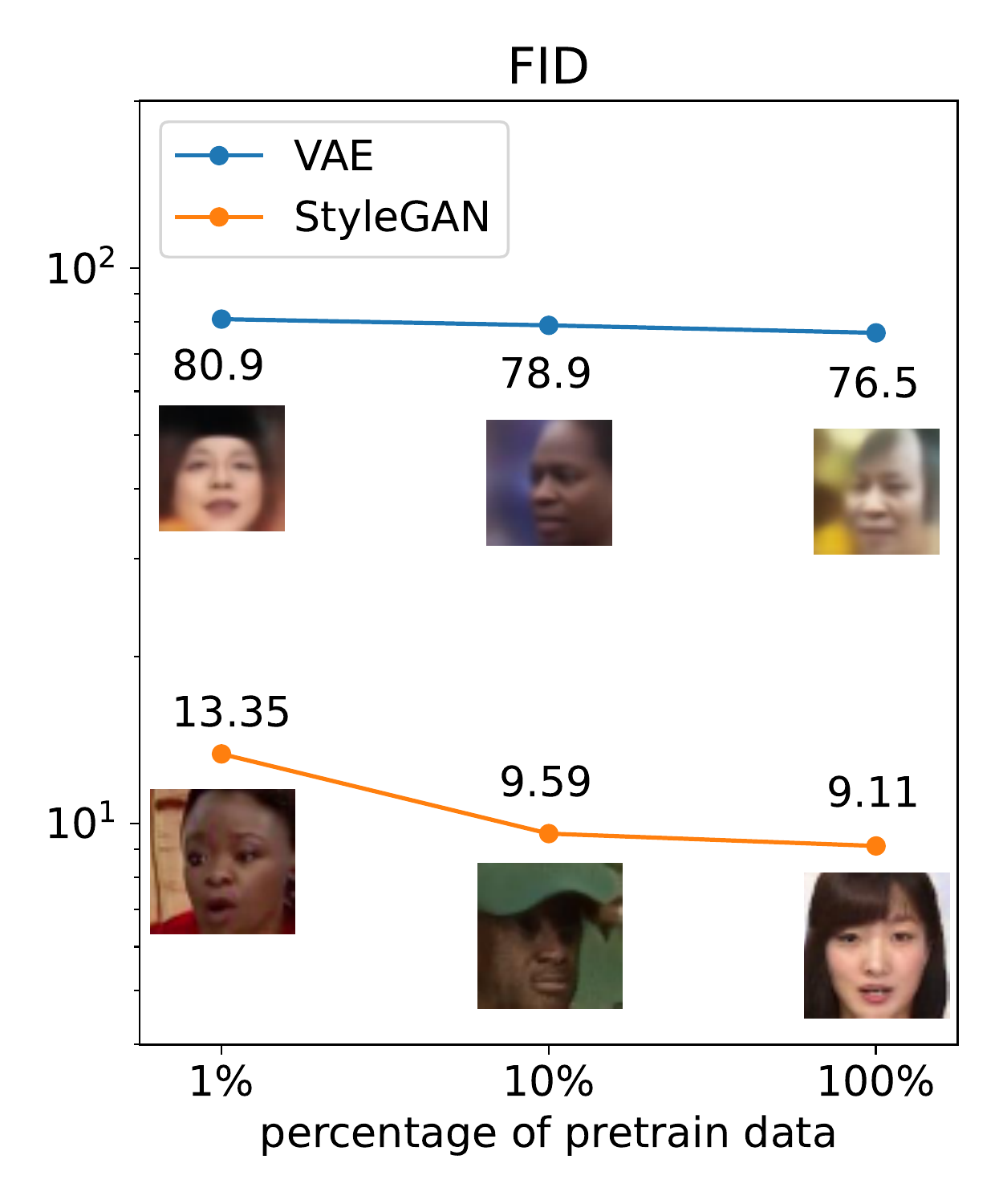}
    \vspace{-0.4cm}
    \caption{FID score, calculated over 100K random pictures from $\mathcal{D}^{prior}$ = \{ AfricanFaceSet-5M $\cup$ AsianFaceSet-3M \}  vs. 100K random samples from VAE or StyleGAN. The score (the less the better) is given w.r.t. the percentage of $\mathcal{D}^{prior}$ used to train these models. We observe that the image generation quality, that FID represents, drops significantly when StyleGAN is provided with more data, while VAE does not scale to larger data sets that well.}\label{fig:vae_vs_stylegan_fid}
    \vspace{-0.5cm}
\end{figure}

\noindent \textbf{Why is AE/VAE less suitable than StyleGAN when fine-tuned for face recognition?} We see two possible reasons why our procedure yields better results. First, StyleGAN-based architectures are tailored to be the state-of-the-art of face generation, which leads to the possible assumption that encoder + StyleGAN pipeline is capable of saving more useful information about the face features than e.g. VAE. Second, we observe that the quality of StyleGAN generations significantly improves when it is trained on larger amount of data, while for VAE it is not the case (6\% over 32\% improvement when 100x more data given -- see Fig.~\ref{fig:vae_vs_stylegan_fid}). This scalability issue is also highlighted by the fact that AE/VAE, trained on the 100\% of $\mathcal{D}^{prior}$, perform for the face recognition task on par with StyleGAN, trained on only 1\% of $\mathcal{D}^{prior}$ (see the Tables~\ref{table:vae} and~\ref{table:rb_webface_wrt_prior_dset_size}).

\noindent \textbf{Is it helpful to draw samples from StyleGAN instead of training an encoder for it?} In this experiment, we used our trained StyleGAN to generate synthetic faces and add them to the training set. We first infer StyleGAN latent codes for all African and Asian samples in our training set (BUPT-BalancedFace). Since face recognition training always requires labeled images, we generate samples using latent codes closer to the inverted training images. In particular, we take random pairs $(I_1, I_2)$ of images of the same ethnic group (either African or Asian) and generate an interpolated sample by calculating convex combinations of their StyleGAN latent codes: $I_\mathrm{comb} = \lambda \cdot f'_{\theta, \omega}(I_1) + (1 - \lambda) \cdot f'_{\theta, \omega}(I_2),\,\, \lambda \sim U[0, 1]$. 
The ground-truth label (always required for face recognition training) is constructed as a \textit{two-hot} vector $(0, \dots, 0, 1 - \lambda, 0, \dots, 0, \lambda, 0, \dots, 0)$ (as influenced by MixUp~\cite{zhang2017mixup}).
In total, we augment the training set with an equal amount of synthetic interpolations (1.3 M), obtained from randomly drawn African-African and Asian-Asian pairs. The results in Table.~\ref{table:synthetic_data_exp} show that the added interpolations, despite the created imbalance and increased variability in the data, do not yield a comparable performance increase. 
In addition, we show that increasing the number of interpolations doesn't help, while pretraining scheme benefits from more pretraining data (see Table~\ref{table:rb_webface_wrt_prior_dset_size}).

\begin{table}
    \centering
\resizebox{\linewidth}{!}{
    \begin{tabular}{p{2.2cm}|cccccccc|}
        & \multicolumn{8}{c|}{RB-WebFace}  \\
        & \multicolumn{4}{c|}{TPR @ FPR=$10^{-3}$ ↑}  & \multicolumn{4}{c|}{TPR @ FPR=$10^{-4}$ ↑} \\ & Cauc.         & Afr.           & Asian             & \multicolumn{1}{c|}{Indian}  & Cauc.         & Afr.           & Asian             & \multicolumn{1}{c|}{Indian}  \\ \hline
        ArcFace R-50                    &    89.86    &    86.73    &    94.31    &    \multicolumn{1}{c|}{93.82}   &    81.48    &    76.80    &    87.47    &     \multicolumn{1}{c|}{86.89}   \\
        \textcolor{pretrainingclr}{+ $\mathcal{D}^{prior}$}  & \textbf{93.54}    &    \textbf{91.51}    &    \textbf{96.51}    &    \multicolumn{1}{c|}{\textbf{96.17}}   & \textbf{86.84}   & \textbf{83.74} &  \textbf{91.74}  & \textbf{91.19}  \\ \hline
        \textcolor{interpolationsclr}{+ 1.3M interps}    &  91.87 & 89.07 & 96.08 & \multicolumn{1}{c|}{95.55}   & 84.65  & 79.97 & 91.13  & 90.30   \\ 
        \textcolor{interpolationsclr}{+ 6M interps}    & 89.48  & 86.85 & 94.92 & \multicolumn{1}{c|}{94.22}   & 80.95  & 77.00 & 88.75  &  87.94  \\ 
        \hline
\end{tabular}
    }
    \caption{In this experiment, instead of pretraining our model as an encoder for StyleGAN, we use the StyleGAN and the trained encoder to draw samples from the learned face distribution and add them to our dataset. The results indicate that  \textcolor{interpolationsclr}{interpolation-based procedure}, despite being more straightforward than the proposed \textcolor{pretrainingclr}{pretraining-based scheme}, is actually less efficient.} \label{table:synthetic_data_exp}
    \vspace{-0.4cm}
\end{table}

We provide additional ablations and comparisons in the Appendix.
 \section{Conclusions}

As an increasing number of modern computer vision methods become production-ready, new challenges regarding their final use are posed. In this work, we present a solution for improving the quality of facial recognition by pretraining the face image encoder with unlabeled data using StyleGAN and encoder. Furthermore, we additionally release two training datasets for unsupervised pretraining and one large-scale protocol for bias estimation. We show that we are able to tune the performance on different ethnicities by altering the composition of the unlabeled prior datasets.
We hope that the released datasets, together with the protocol, will drive forward the research on mitigating the face recognition biases using self-supervised approaches.

There are several directions for possible future work that we foresee. First, integrating the two steps of pretraining and fine-tuning into one would help to avoid forgetting of the pretrained weights. Second, publicly available datasets often don't possess enough variability (both in terms of capture conditions and demographics)~\cite{Hazirbas2021towards} and typically feature only low-resolution images which might be a blocking factor for the fairness research~\cite{Wang2020mitigating,Knoche2021image}. Accordingly, the results on the newly assembled RB-WebFace benchmark can be analyzed more thoroughly, which can bring new insights about the factors positively affecting bias mitigation in real-world scenarios. Finally, various architectures typically used for self-supervised learning (such as transformers or highly scalable generative models~\cite{sauer2022stylegan}), might allow for the construction of more flexible methods due to heterogeneous inputs and outputs.

\noindent \textbf{Legal concerns.} The collected unlabeled data was collected anonymously in accordance with Standard YouTube License and CC YouTube License and does not contain personally identifiable information (PII). The data is released only in the form of links to YouTube videos and the corresponding timestamps, fully following good practice of similar data collection~\cite{chung2018voxceleb2,ephrat2018looking,li2020celeb} and with the data subjects protection as per Art.~14~5(b) GDPR law.

\noindent \textbf{Acknowledgments.} We gratefully acknowledge the support of this research by a TUM-IAS Hans Fischer Senior Fellowship, the ERC Starting Grant Scan2CAD (804724) and the Horizon Europe vera.ai project (101070093). We also thank Yawar Siddiqui for the helpful technical advice on StyleGAN part, Angela Dai for the video voiceover, and Dmitriy Karfagenskiy for the proofreading and corrections.
 
{\small
\bibliographystyle{ieee_fullname}
\bibliography{egbib}
}
\cleardoublepage

\begin{appendix}
\section{Data collection}
\label{sec:data_collection}

To scrape AfricanFaceSet-5M and AsianFaceSet-3M, we followed the same pipeline. First, a list of channels with predominantly African \textit{(List 1)} and Asian \textit{(List 2)} demographics is provided (see below). List of IDs of all videos released in each of the channels is collected via \href{https://github.com/dermasmid/scrapetube}{scrapetube} library. Typical number of videos in a channel is 50K--150K. Afterwards, the videos are downloaded one-by-one in the highest available quality with a limit of 20 min per video via \href{https://github.com/pytube/pytube}{pytube} in 36 parallel threads. The time limit is specified in the video URL, which allows not to download excessive parts of videos. One frame per every $P=5$ is extracted via \href{https://ffmpeg.org/}{ffmpeg}, and faces are extracted using MTCNN~\cite{Zhang2016joint} (\href{https://github.com/TropComplique/mtcnn-pytorch}{mtcnn-pytorch} implementation) and are aligned by landmarks via \texttt{cv2.warpAffine} from \href{https://opencv.org/}{OpenCV}. To ensure the correctness of the MTCNN face detector, we impose constraints in terms of the minimal face size in the original image ($100 \times 100$ px) and value of 0.9 set for all the detector thresholds. Example pictures are provided in Fig.~\ref{fig:africanfaceset} and Fig.~\ref{fig:asianfaceset}.

\begin{figure}[h!]
    \centering
    \begin{subfigure}[b]{0.49\linewidth}
        \centering
        \includegraphics[trim={0 0 0 0},clip,width=\linewidth]{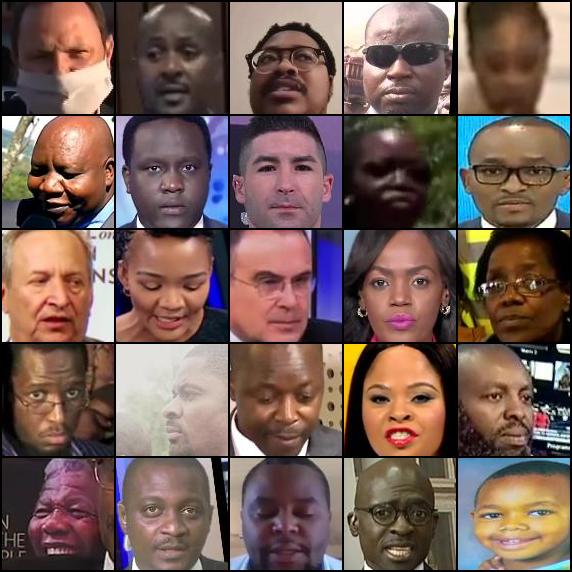}
        \caption{Examples from AfricanFaceSet-5M}
        \label{fig:africanfaceset}
    \end{subfigure}
    \hfill
    \begin{subfigure}[b]{0.49\linewidth}
        \centering
        \includegraphics[trim={0 0 0 0},clip,width=\linewidth]{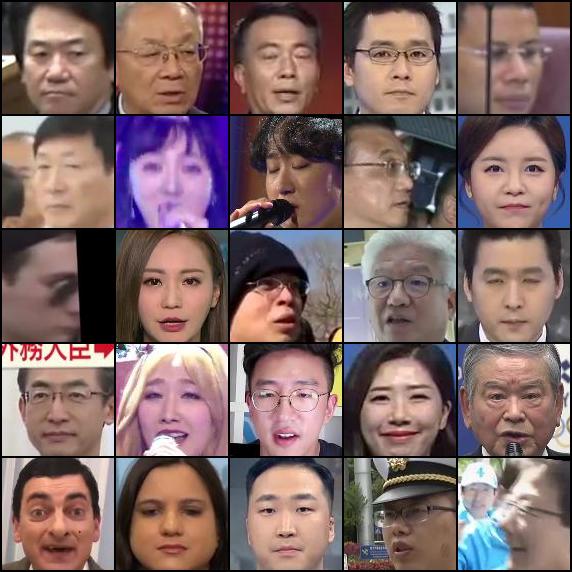}
        \caption{Examples from AsianFaceSet-3M}
        \label{fig:asianfaceset}
    \end{subfigure}
    \caption{Random samples from the collected AfricanFaceSet-5M and AsianFaceSet-3M datasets.}
\end{figure}

YouTube channels used for the collection of AfricanFaceSet-5M \textit{(List 1)}:

\begin{footnotesize}
\begin{itemize}
    \item Africa24 \url{https://www.youtube.com/user/Africa24}
    \item African Glitz \url{https://www.youtube.com/c/AfricanGlitzTV}
    \item Arise News \url{https://www.youtube.com/c/AriseNewsChannel}
    \item BBC News Africa \url{https://www.youtube.com/c/BBCAfrica}
    \item Best African TV \url{https://www.youtube.com/channel/UCz2lohZJpkfOkvXVyu_6wng}
    \item CGTN Africa \url{https://www.youtube.com/c/cgtnafrica}
    \item Channels Television \url{https://www.youtube.com/c/ChannelsTelevision}
    \item DStv \url{https://www.youtube.com/dstv}
    \item eNCA \url{https://www.youtube.com/c/encanews}
    \item Eye Witness News \url{https://www.youtube.com/c/EyeWitnessNewsBahamas}
    \item Guardian Nigeria \url{https://www.youtube.com/c/GuardianNigeriaOfficial}
    \item Legit TV \url{https://www.youtube.com/c/LegitTV}
    \item News Central TV \url{https://www.youtube.com/c/NewsCentralTVafrica}
    \item Newzroom Africa \url{https://www.youtube.com/channel/UCQMML3hAsx-Mz9j9ZN0tThQ}
    \item One Africa TV \url{https://www.youtube.com/c/OneAfricaTelevision}
    \item Plus TV Africa \url{https://www.youtube.com/c/PlusTVAfrica}
    \item Roots TV \url{https://www.youtube.com/c/RootsTVCommunity}
    \item SABC News \url{https://www.youtube.com/sabcnews}
    \item TVC News Nigeria \url{https://www.youtube.com/c/tvcnewsnigeria}
    \item Voice TV Nigeria \url{https://www.youtube.com/c/VoicetvNigeria}
    \item The Walk \url{https://www.youtube.com/c/TheWalkk} (without the 20 min video limit, since the channel contains long walking tours over cities)
    \item Kenya Citizen TV \url{https://www.youtube.com/c/kenyacitizentv}
    \item Africa News \url{https://www.youtube.com/c/africanews}
    \item African Tigress \url{https://www.youtube.com/c/AFRICANTIGRESS}
\end{itemize}
\end{footnotesize}

YouTube channels used for the collection of AsianFaceSet-3M \textit{(List 2)}:

\begin{footnotesize}
\begin{CJK*}{UTF8}{gbsn}
\begin{itemize}
    \item Asian Boss \url{https://www.youtube.com/c/AsianBoss}
    \item CCTV Video News Agency \url{https://www.youtube.com/c/CCTVVideoNewsAgency}
    \item China Daily 中国日报~\url{https://www.youtube.com/channel/UCahujLjSL34EPNxtwKRi_vg}
    \item China Live 直播中国~\url{https://www.youtube.com/c/chinanews}
    \item China Matters \url{https://www.youtube.com/c/ChinaMatters}
    \item CNA \url{https://www.youtube.com/user/channelnewsasia}
    \item Discovery Channel Southeast Asia \url{https://www.youtube.com/c/DiscoveryChannelSEAsia}
    \item New China TV \url{https://www.youtube.com/c/ChinaViewTV}
    \item Nikkei Asia \url{https://www.youtube.com/user/NikkeiAsia}
    \item South China Morning Post \url{https://www.youtube.com/c/SouthChinaMorningPost}
    \item Tencent Video \url{https://www.youtube.com/channel/UCQatgKoA7lylp_UzvsLCgcw}
    \item Top Korean News \url{https://www.youtube.com/c/TopKoreanNews}
    \item ANNnewsCH \url{https://www.youtube.com/user/ANNnewsCH}
    \item Arirang News \url{https://www.youtube.com/c/ArirangCoKrArirangNEWS}
    \item Ask Japanese \url{https://www.youtube.com/c/AskJapanese}
    \item CCTV Video News Agency \url{https://www.youtube.com/c/CCTVVideoNewsAgency}
    \item CCTV中国中央电视台~\url{https://www.youtube.com/c/cctv}
    \item CCTV今日说法官方频道~\url{https://www.youtube.com/user/jinrishuofa}
    \item CCTV挑战不可能官方频道 \newline \url{https://www.youtube.com/channel/UC3HLhJGcc_0Vse2UncGnxcQ}
    \item CCTV春晚~\url{https://www.youtube.com/c/cctvgala}
    \item CCTV电视剧~\url{https://www.youtube.com/channel/UC7Vl0YiY0rDlovqcCFN4yTA}
    \item CCTV社会与法~\url{https://www.youtube.com/c/Internationalcntv}
    \item CCTV科教~\url{https://www.youtube.com/user/kejiaotv}
    \item CCTV纪录~\url{https://www.youtube.com/user/documentarycntv}
    \item DKDKTV \url{https://www.youtube.com/c/DKDKTV}
    \item Hi China \url{https://www.youtube.com/c/CCTVcomInternational}
    \item KBS WORLD TV \url{https://www.youtube.com/c/kbsworldtv}
    \item KOREA NOW \url{https://www.youtube.com/c/KOREANOWyna}
    \item Live Japan \url{https://www.youtube.com/channel/UCW879NMJHIvKspfOg3H8OsQ}
    \item NHK WORLD-JAPAN \url{https://www.youtube.com/c/NHKWORLDJAPAN}
    \item Nippon TV News 24 Japan \url{https://www.youtube.com/c/NipponTVNews24Japan}
    \item ShanghaiEye~魔都眼~\url{https://www.youtube.com/c/Kankanewsbilingual}
    \item The Japan Times \url{https://www.youtube.com/user/thejapantimes}
    \item The Thaiger \url{https://www.youtube.com/c/TheThaiger}
    \item Tokyo Explorer \url{https://www.youtube.com/c/TokyoExplorer}
    \item VisitSeoul TV \url{https://www.youtube.com/c/VisitSeoulTV}
    \item Walk East \url{https://www.youtube.com/c/WalkEast}
    \item 無綫新聞~TVB NEWS official\newline \url{https://www.youtube.com/channel/UC_ifDTtFAcsj-wJ5JfM27CQ}
\end{itemize}
\end{CJK*}
\end{footnotesize}

\begin{table*}[]
\bgroup
\resizebox{\linewidth}{!}{
\begin{tabular}{p{0.2\linewidth}|cccccc}
Dataset name & \# people & \# pos pairs & \# neg pairs & \# subgroups & neg pairs not by facerec & preferred \linebreak protocol \\ \hline
IJB-A        & 500       & $\le$ 23K          & 100K           & 1                                & \cmark    & ROC curve                              \\
DemogPairs   & 600       & 91K          & 19M          & 6                                & \cmark    & ROC curve                              \\
BFW          & 800       & 240K         & 681K         & 8                                & \cmark    & ROC curve                              \\
RFW          & 12K       & 12K          & 12K          & 4                                & \xmark    & LFW-like                               \\ \hline
RB-WebFace   & \textbf{72K}       & \textbf{360K}         & \textbf{648M}         & 4                                & \cmark    & ROC curve                             
\end{tabular}
}\newline
\egroup
\caption{Overview of the existing publicly available datasets of pairs used to evaluate face recognition accuracy. Since the appearance of LFW~\cite{Huang2008labeled}, many test sets consisting of the same number positive (same person) pairs and negative (similar people) pairs have been proposed. The RFW dataset~\cite{Wang2019racial} is compiled from MS-Celeb-1M~\cite{Guo2016celeb} in a similar fashion for the purpose of fairness estimation of a trained face recognizer and considered the standard benchmark for fairness. We propose a new testing set for the fairness estimation -- \textit{RB-WebFace} -- comprising a partition of recently released WebFace-42M, which addresses two issues of RFW. First, we use all negative pairs instead of their subset selected by a pretrained face recognition network that can be potentially introduce selection bias. Second, the dataset contains much larger number of pairs. As we show, RB-WebFace is also a harder (less saturated) benchmark.}
\label{table:test-datasets}
\end{table*}

\begin{figure}[h!]
    \centering
    \begin{subfigure}[b]{0.45\textwidth}
        \centering
        \includegraphics[trim={0 0 0 0},clip,width=\textwidth]{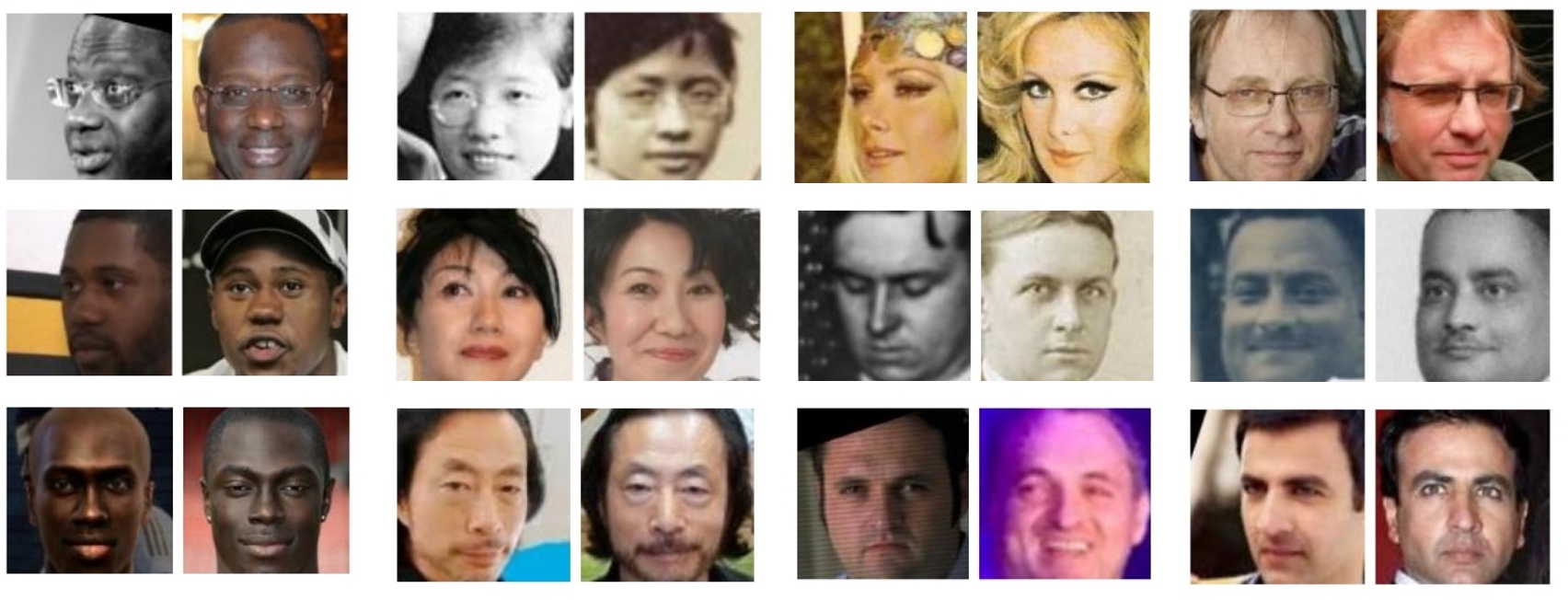}
        \caption{RFW positive pairs examples}
    \end{subfigure}
    \hfill
    \begin{subfigure}[b]{0.45\textwidth}
        \centering
        \includegraphics[trim={0 0 0 0},clip,width=\textwidth]{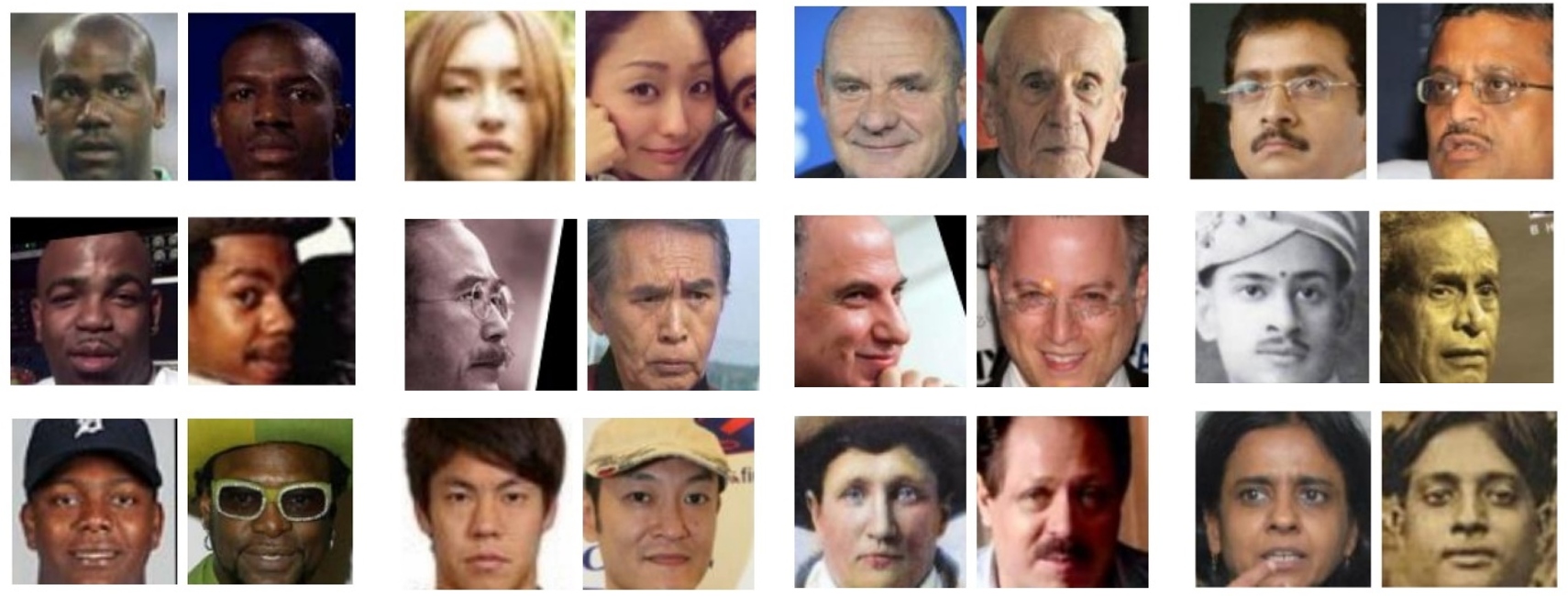}
        \caption{RFW negative pairs examples}
    \end{subfigure}
    
    \centering
    \begin{subfigure}[b]{0.45\textwidth}
        \centering
        \includegraphics[trim={0 0 0 0},clip,width=\textwidth]{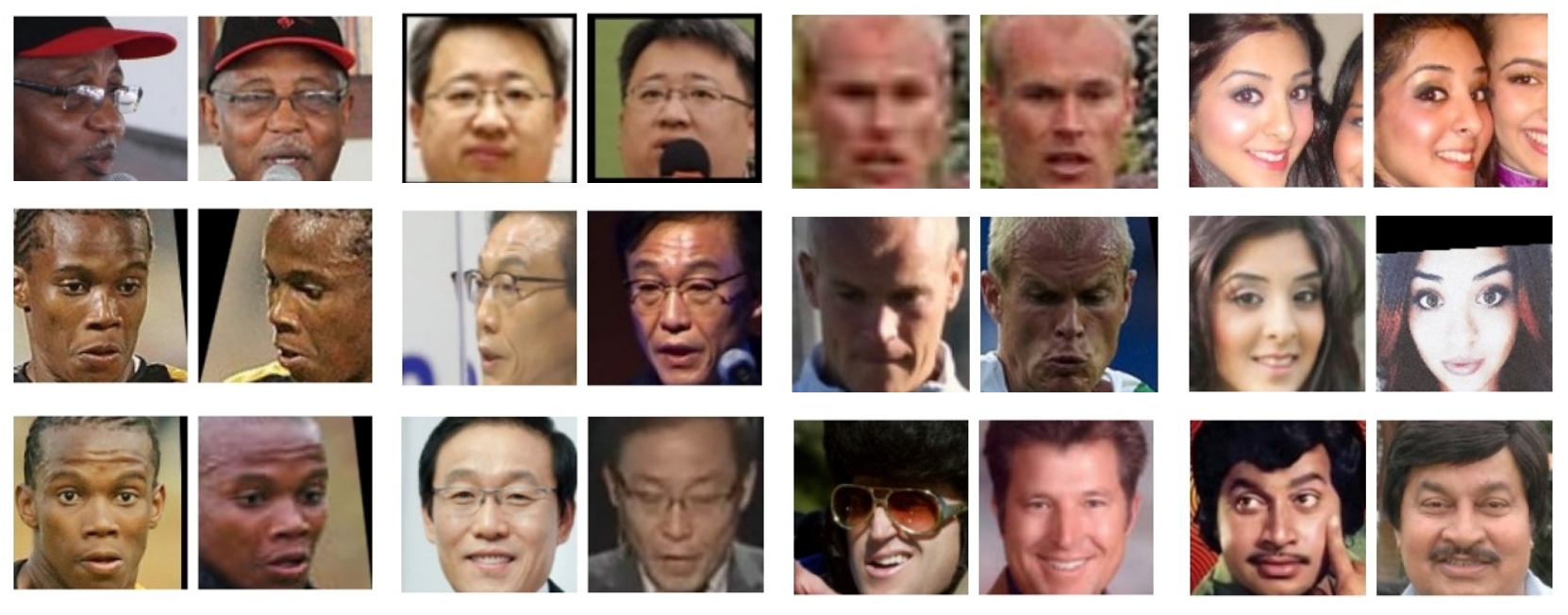}
        \caption{RB-WebFace positive pairs examples}
    \end{subfigure}
    \hfill
    \begin{subfigure}[b]{0.45\textwidth}
        \centering
        \includegraphics[trim={0 0 0 0},clip,width=\textwidth]{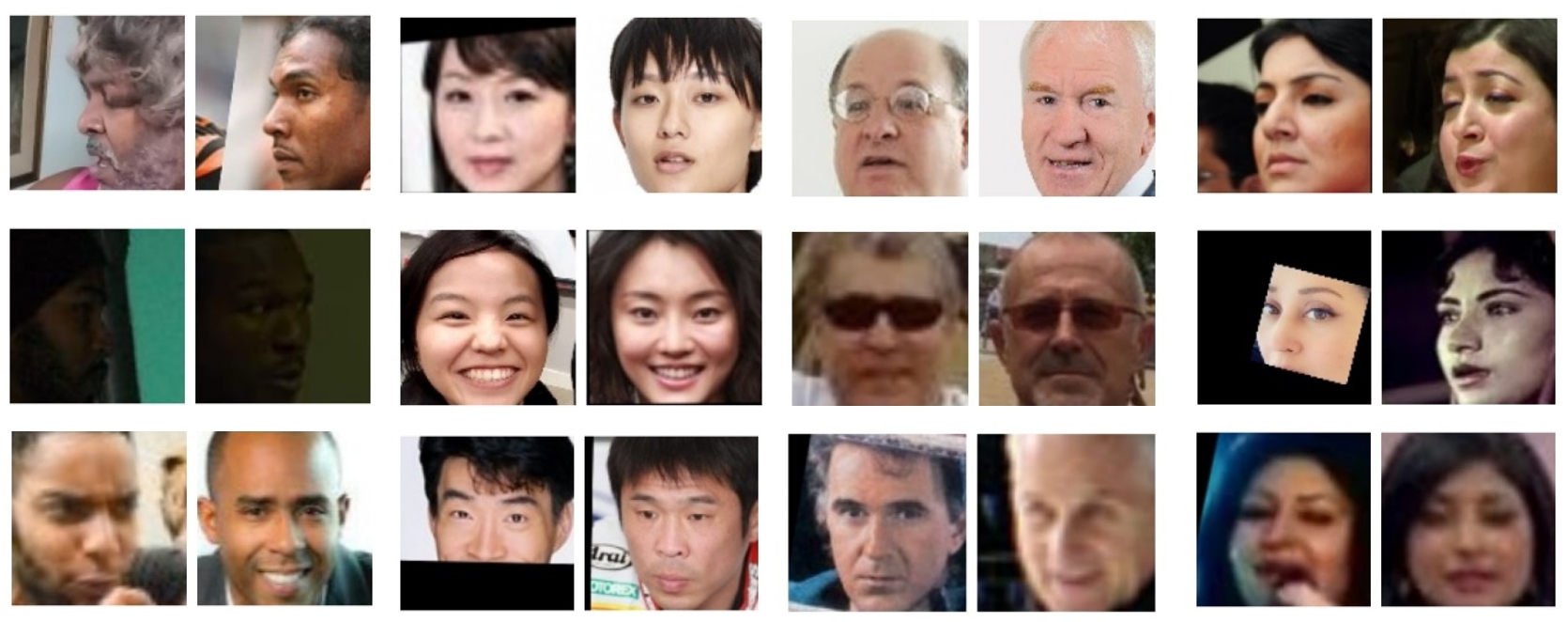}
        \caption{RB-WebFace negative pairs examples}
    \end{subfigure}
    \caption{Examples of positive and negative pairs on RFW and newly assembled RB-WebFace (partition on WebFace-42M).}
    \label{fig:pos_neg_pairs}
\end{figure}

The examples of positive and negative pairs of both publicly available RFW and newly assembled RB-WebFace are shown in Fig.~\ref{fig:pos_neg_pairs}. As shown, both datasets feature challenging pairs, however in RB-WebFace the evaluation protocol is different: for RB-WebFace, we calculate TPR given predefined FPR for a small number of positive pairs and a large number of negative pairs, while for RFW, simple calculation of accuracy is possible, since the number of positive pairs and negative pairs is the same. Using all possible negatives for RB-Webface allows to reduce the potential selection bias, as there is no longer any need to select challenging negatives by a face recognition network.

\section{Additional comparisons}
\label{sec:ablations}

\begin{figure*}[t!]
 \centering
     \begin{subfigure}[b]{0.24\linewidth}
         \centering
         \includegraphics[trim={0 0 0 0},clip,width=\linewidth]{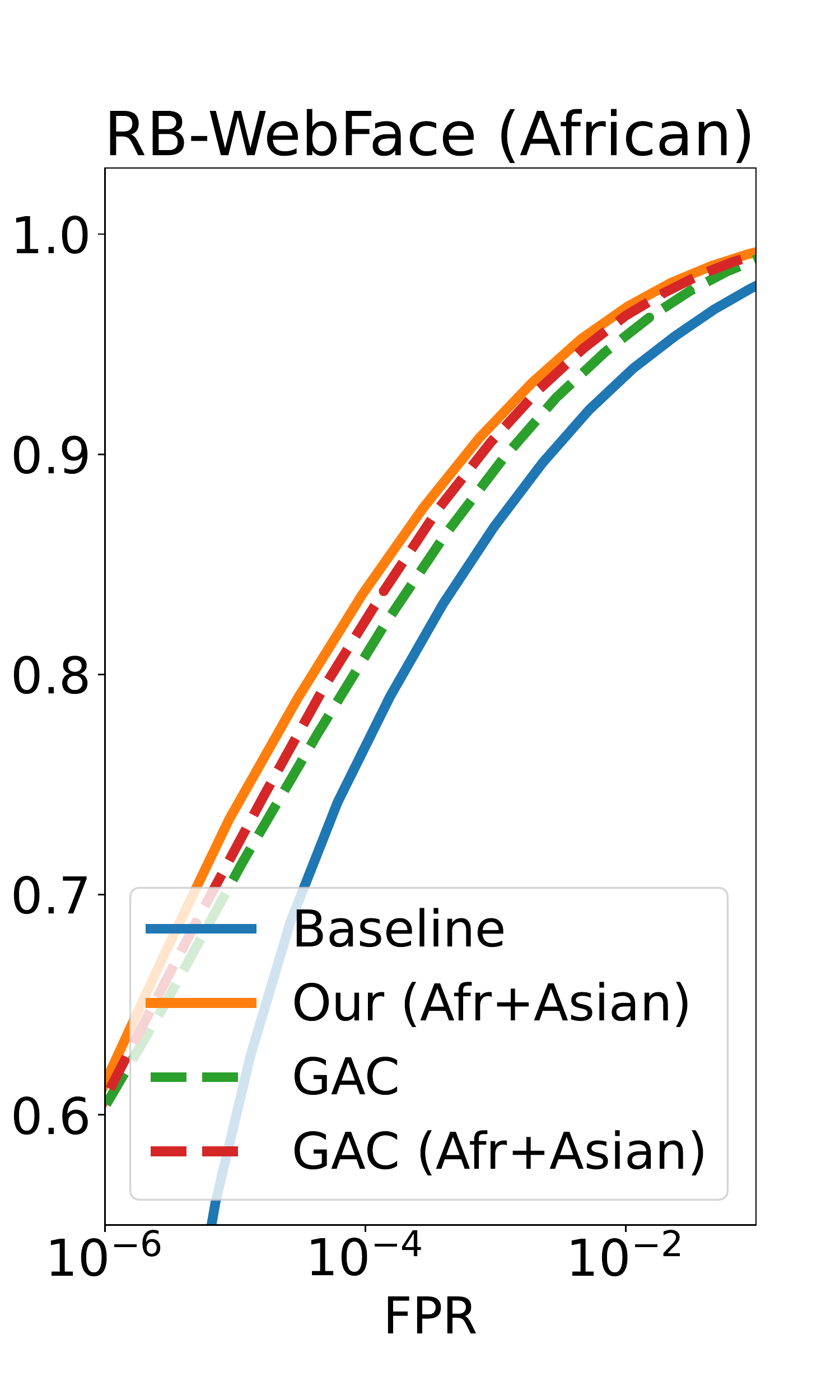}
     \end{subfigure}
     \hfill
     \begin{subfigure}[b]{0.24\linewidth}
         \centering
         \includegraphics[trim={0 0 0 0},clip,width=\linewidth]{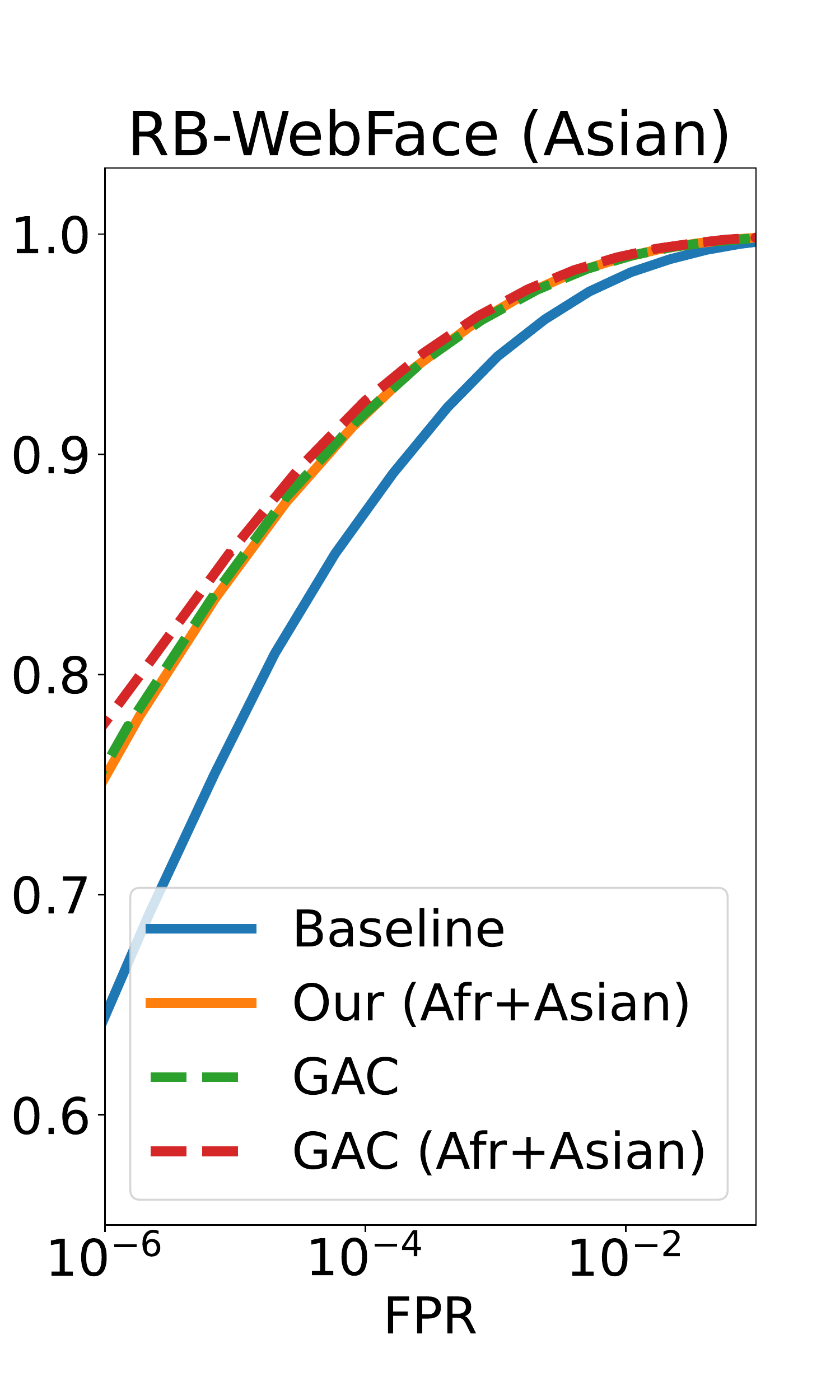}
     \end{subfigure}
     \hfill
     \begin{subfigure}[b]{0.24\linewidth}
         \centering
         \includegraphics[trim={0 0 0 0},clip,width=\linewidth]{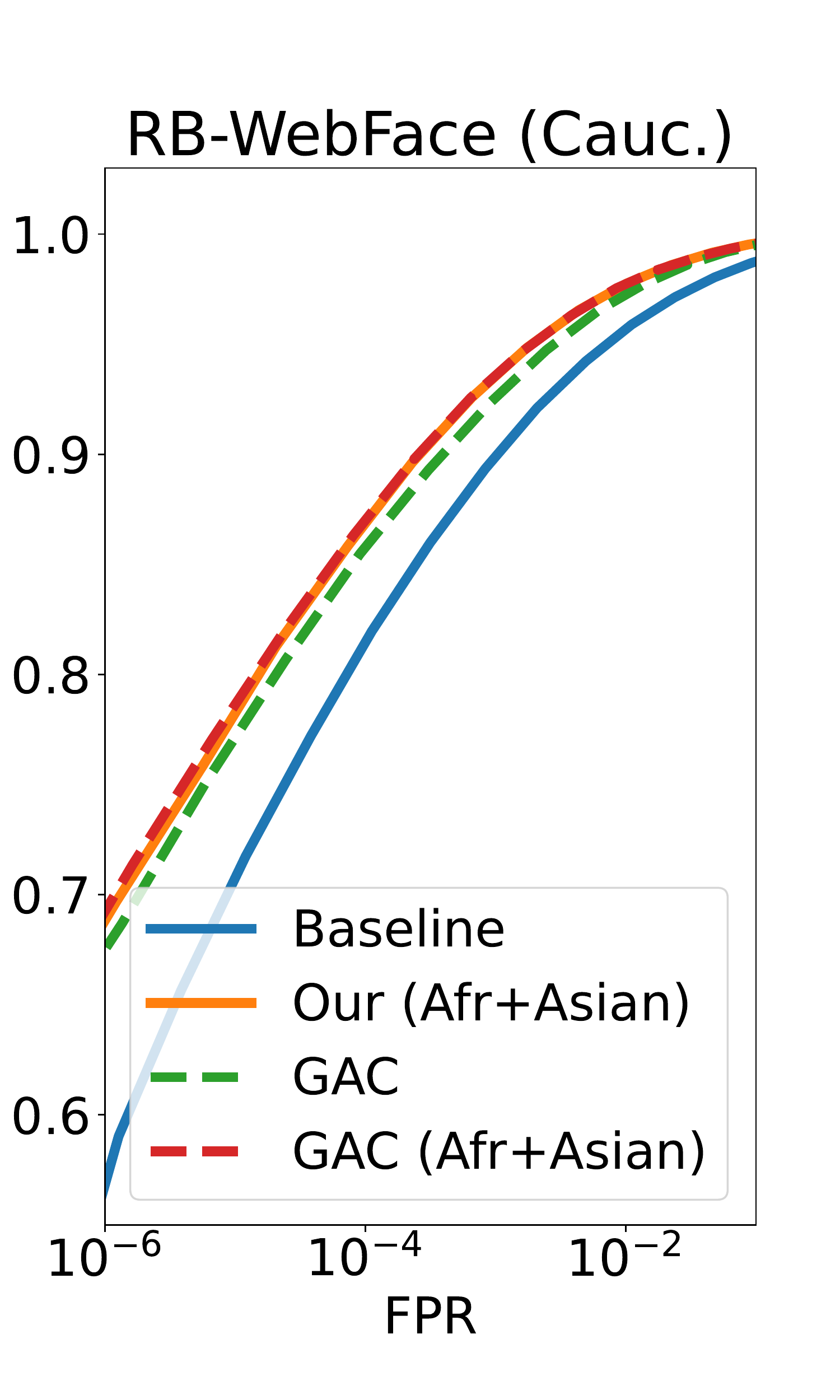}
     \end{subfigure}
     \hfill
     \begin{subfigure}[b]{0.24\linewidth}
         \centering
         \includegraphics[trim={0 0 0 0},clip,width=\linewidth]{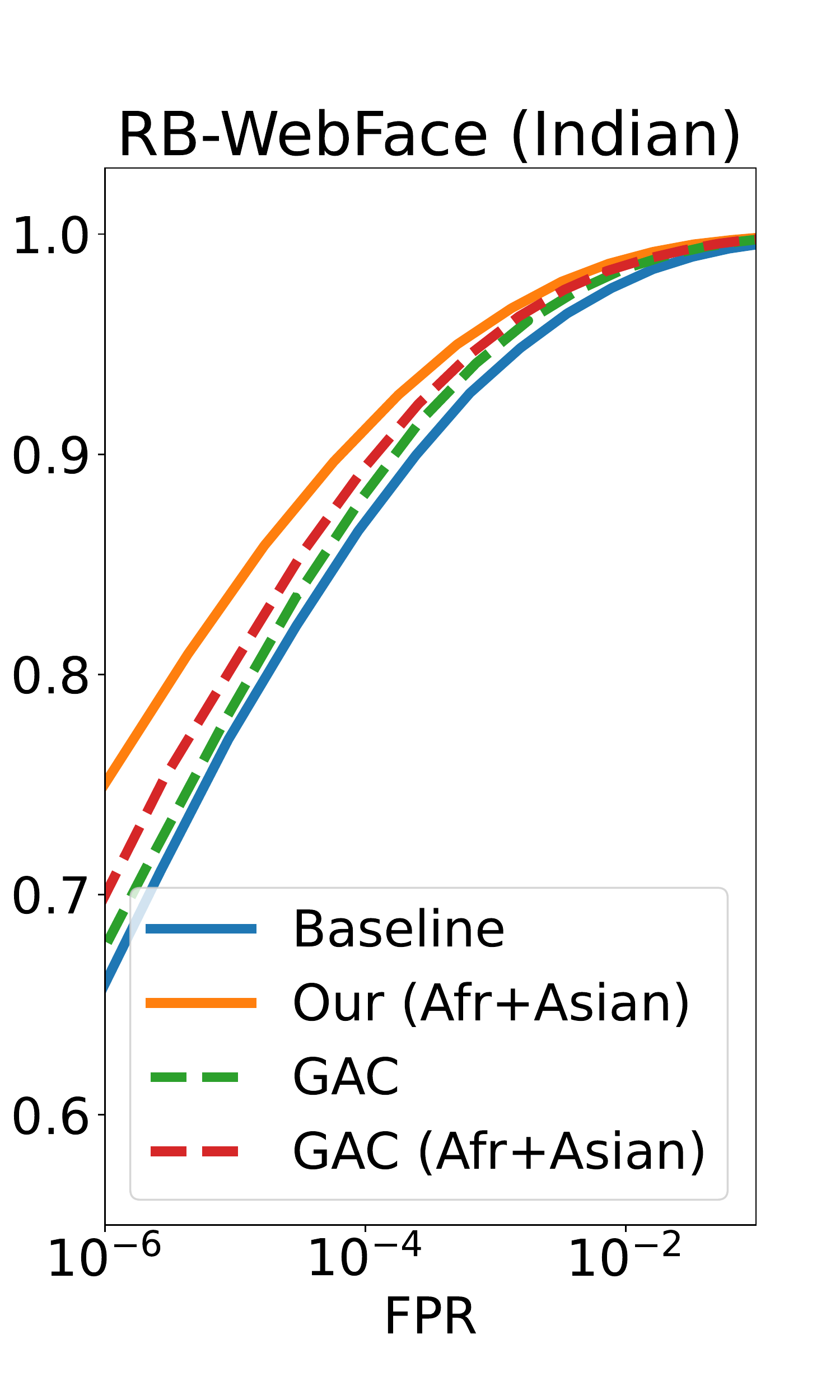}
     \end{subfigure}
 \caption{Comparison of the ROC curves for the methods on the newly assembled RB-WebFace validation set. Similarly to RFW, RB-WebFace consists of positive and negative pairs constructed from the set of samples. In the plot legend, \textit{Ours} refers to our method, while \textit{Baseline} stands for the \textit{ArcFace R-50} baseline. \textit{GAC} denotes \textit{GAC (Ground truth)} method and \textit{GAC (Afr+Asian)} describes its version with the proposed pretraining on $\mathcal{D}^{prior}$. Note the constant increase of TPR for the versions of algorithms enhanced by the proposed pretraining procedure.}
 \label{fig:roc_curves}
\end{figure*}

\textbf{TPR and FPR values at all thresholds.} The comparison in Fig.~\ref{fig:roc_curves} is a graphical representation of quality of the methods on RB-WebFace described in Table~3 in the main paper text. Here we showcase the same data not for the fixed FPR but for all FPR in the form of plots, obtained by sweeping a threshold.

\textbf{Effect of the collected prior datasets vs. using FFHQ.} Here we evaluate whether it makes sense to employ larger and more diverse unlabeled data collections in the pretraining stage by comparing \textcolor{pretrainingclr}{$\mathcal{D}^{prior}$} = AfricanFaceSet $\cup$ AsianFaceSet to \textcolor{pretrainingclr}{$\mathcal{D}^{prior}$} = FFHQ. Relative to baseline, pretraining on FFHQ helps all ethnicities and mostly Caucasian, which is indeed the predominant group in FFHQ.

\begin{table}[h!]
    \vspace{-0.2cm}
    \begin{center}
    \resizebox{1.0\linewidth}{!}{
        \begin{tabular}{l|cccc|cc}
                     & \multicolumn{6}{c}{RFW, accuracy \%, ↑}                    \\ 
                     & Cauc.      & African        & Asian          & Indian  & avg $\uparrow$  & std $\downarrow$      \\ \hline
        ArcFace R-50 & 96.00          & 94.00          & 93.08          & 94.48     &      94.39  & 1.06   \\
        \textcolor{pretrainingclr}{+ $\mathcal{D}^{prior}$ (Afr+Asian)}  & 96.52          & \textbf{95.00} & \textbf{93.90} & \textbf{94.93}  & \textbf{95.09}  & \textbf{0.94} \\
        \textcolor{pretrainingclr}{+ $\mathcal{D}^{prior}$ (FFHQ)}       & \textbf{96.58} & 94.42          & 93.65          & 94.53      & 94.80   & 1.09    \\ \hline
        \end{tabular}
    }
    \end{center}
    \caption{Comparison to pretraining on FFHQ dataset.}
    \label{table:ffhq}
\end{table}

\textbf{Comparison of the StyleGAN encoder architectures.} We provide the ablation over encoder training strategies in Table~\ref{table:encoders}. 
There's no specific strategy that yields the best result across all groups, but by avg and std, pSp is the best-performing choice of architecture.

\begin{table}[h!]
    \vspace{-0.15cm}
    \begin{center}
    \resizebox{\linewidth}{!}{
        \begin{tabular}{l|cccc|cc}
        Method to train   & \multicolumn{6}{c}{RFW, accuracy \%, ↑}       \\ the R-50 encoder & Cauc. & African   & Asian     & Indian  & avg $\uparrow$ & std $\downarrow$  \\ \hline
        \textcolor{pretrainingclr}{pSp} \cite{Richardson2021encoding}           & 96.52     & \textbf{95.00}     & 93.90     & 94.93  & \textbf{95.09} & \textbf{0.94}   \\
        \textcolor{pretrainingclr}{e4e} \cite{Tov2021designing}          & 96.40     & 94.08     & \textbf{94.10}     & \textbf{95.05} & 94.91 & 0.95 \\
        \textcolor{pretrainingclr}{ReStyle} \cite{Alaluf2021restyle} & \textbf{96.67} &  94.43         & 93.83    & 94.80    & 94.93  & 1.06      \\ \hline
        \end{tabular}
    }
    \end{center}
    \vspace{-0.3cm}
    \caption{Comparison of StyleGAN encoders to use in Stage 2 and 3 (\textcolor{pretrainingclr}{Afr+Asian}). ReStyle is based on cascaded prediction and, in this experiment, iterates through pSp base architecture three times per pass.}
    \label{table:encoders}
\end{table}

\textbf{Application for gender classification}. The demonstrated encoder-based pretraining technique is also applicable to other tasks. To show that, we conduct a simple experiment where the pSp R-34 encoder, pretrained in Stage 2, is fine-tuned for gender classification, not face recognition task. As a labeled dataset, we take \href{https://www.kaggle.com/datasets/ashishjangra27/gender-recognition-200k-images-celeba?resource=download}{Kaggle 200K gender recognition from CelebA} dataset, and fine-tune the encoder on it with binary cross-entropy loss. Just like for our primary downstream task, the results indicate that the quality boost is especially prominent for a limited amount of labeled data. When trained on 1\% of the labeled dataset, \textbf{94.42\%} accuracy is achieved with our pretraining and \textbf{75.17\%} without it. For 10\% of the labeled dataset, \textbf{97.04\%} accuracy with our pretraining vs \textbf{93.47\%} without is achieved. For the full dataset, the quality difference was saturated. In this experiment, we freeze the encoder for the first 8 epochs when training on 1\% of the labeled dataset (both w/ and w/o pretraining) to avoid SGD convergence issues.

\textbf{Prior datasets filtering}. 
We found that applying strict ethnicity filtering via consensus-based classifier (see Subsec.~\ref{subsec:rb_webface}) on AfricanFaceSet and AsianFaceSet removes around 30\% of the collected faces. Unlike the case when no filtering is performed (Table 2 in the main paper text), pretraining on the filtered data results in more evident same-race improvement (i.e., pretraining on African helps more on RFW-African benchmark, and pretraining on Asian helps more on RFW-Asian) -- see Table~\ref{table:filtered}. 

\begin{table}[h]
\vspace{-0.1cm}
\begin{center}
\resizebox{1.0\linewidth}{!}{
\begin{tabular}{l|cccc}
                                                                  & \multicolumn{4}{c}{RFW, accuracy \%, ↑}                          \\ 
                                                                  & Cauc.      & African        & Asian          & Indian               \\ \hline
Baseline (ArcFace R-50)                            &       96.00                     & 94.00          & 93.08          & 94.48        \\
Baseline \textcolor{pretrainingclr}{+ $\mathcal{D}^{prior}$ (African-\textbf{F})}     &     96.10         & \textbf{94.93}          & 93.70          & 95.27        \\
Baseline \textcolor{pretrainingclr}{+ $\mathcal{D}^{prior}$ (Asian-\textbf{F})}       &     96.70          & 94.53          & \textbf{94.23}          & 94.75                   \\ \hline
\end{tabular}
}\newline
\caption{Comparison on RFW with filtered (\textcolor{pretrainingclr}{-\textbf{F}}) prior datasets.}
\label{table:filtered}
\end{center}
\end{table}

Even with the filtering applied, certain improvement for one ethnicity can be observed even after pretraining on another ethnicity (e.g., \textcolor{pretrainingclr}{Asian-\textbf{F}} also helps on African). 
The improvement can probably be attributed to transfer learning of general face features/conditions/geometry  independent of the subject ethnicity.
For instance, FFHQ pretraining, predominantly Caucasian, also aids in recognizing other ethnicities (see Table~\ref{table:ffhq}).

\section{Implementation details}
\label{sec:implementation}

\subsection{RB-WebFace}
\label{subsec:rb_webface}

Here we provide additional information about the construction of the RB-WebFace protocol, which is done in several stages. First, images from WebFace-42M are processed by an ethnic group classifier pretrained on BUPT-BalancedFace that makes the initial judgment of whether the person belongs to the African, East Asian, Indian, or Caucasian group. Since WebFace-42M contains several images per each of its 2M people, we apply the consensus algorithm to make the classifier's decision more confident about the person's ethnic group. Specifically, we consider the person belonging to the ethnic group $E,\, E \in \{1,.., 4\}$, if there are at least 14 photos of this person in the dataset, and the ethnic group classifier predicts the group $E$ for at least 80\% of their photos (not more than 20 random photos of the person are considered). Subsequently, $N$ people from each group are selected and $M$ positive pairs are constructed from them. A set of negative pairs is constructed as a compilation of all possible distinct pairs of $N$ pictures (1 random image of each person). We used the maximal possible value for $N=18000$ (the lowest number of people across 4 ethnic groups, for which the race classifier was sure about the race). For each person, five positive pairs are selected, resulting in 90 K positive and $\sim$162 M negative pairs per ethnic group. Since a pretrained face recognition network can potentially induce bias in the selection of negative pairs, we deliberately make use of all the possible $\mathcal{O}({N^2})$ negative pairs.

\subsection{Training procedure}
\label{subsec:training_procedure}

\indent This section reveals a number of implementation details not covered in the main paper text.

\begin{figure}[h!]
\centering
\centering
    \begin{subfigure}[b]{0.48\linewidth}
        \centering
        \includegraphics[trim={0 0 0 0},clip,width=\linewidth]{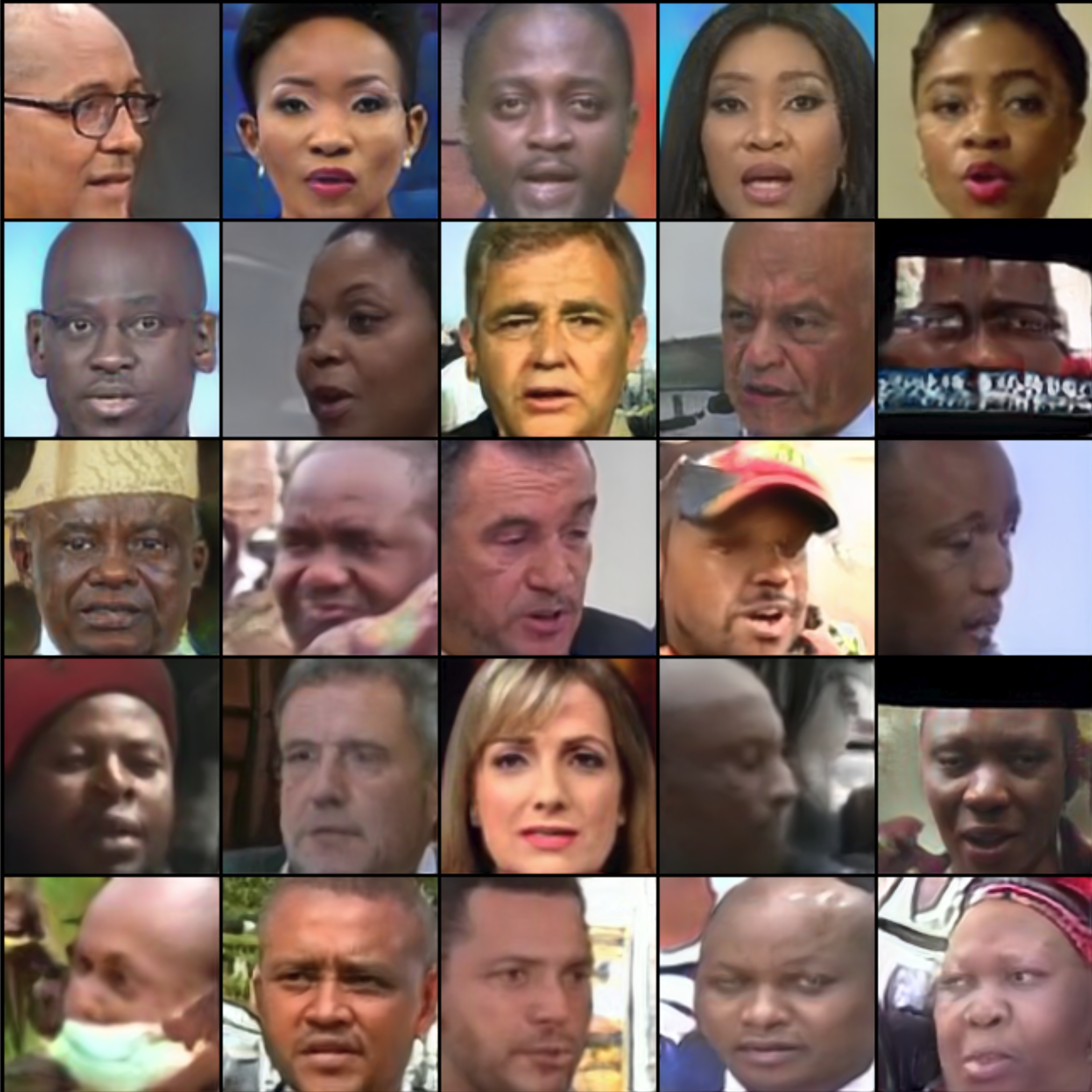}
        \caption{Generations for StyleGAN trained on AfricanFaceSet-5M}
        \label{fig:africanfaceset}
    \end{subfigure}
    \hfill
    \begin{subfigure}[b]{0.48\linewidth}
        \centering
\includegraphics[trim={0 0 0 0},clip,width=\linewidth]{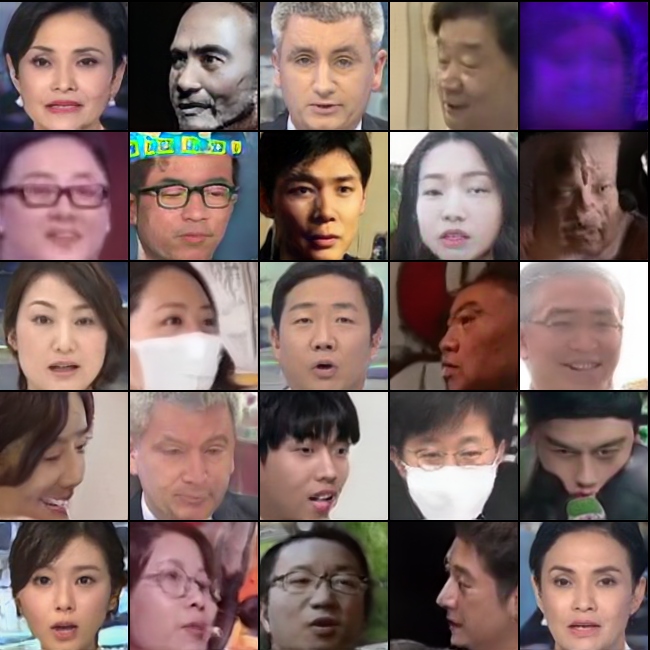}
        \caption{Generations for StyleGAN trained on AsianFaceSet-3M}
        \label{fig:asianfaceset}
    \end{subfigure}
    \caption{Random generations by a StyleGAN trained on either AfricanFaceSet-5M or AsianFaceSet-3M.}
\end{figure}

For the first pretraining stage (StyleGAN2-ADA training), we used the \linebreak \href{https://github.com/nihalsid/stylegan2-ada-lightning}{stylegan2-ada-lightning} implementation and trained it with the following hyperparameters:

\begin{table}[h!]
\resizebox{\linewidth}{!}{
\begin{tabular}{c|c|c|c}
latent dim & \# layers (z $\rightarrow$ w) & G lr  & D lr     \\ \hline
512        & 8                             & 0.002 & 0.00235 \\
\multicolumn{4}{c}{} \\
$\lambda_{gp}$ & $\lambda_{plp}$ & ada\_start\_p & ada\_target \\ \hline
4.0            & 2.0            & 0.0           & 0.6    \\
\end{tabular}
}
\end{table}

\noindent The number of samples seen during training is set to 8 million, which roughly corresponds to the observed number of iterations when FID reconstruction score stops decreasing during fitting. The choice of $\lambda_{gp} = 4$ and an 8-layer mapping network is relatively unconventional and proved best in our setting. The resolution of output images was set to $128 \times 128$. Training was performed on 4 NVIDIA RTX 2080 Ti GPUs with 11 GB memory size each, with minibatch size of 32 and without mixed precision.

For the second pretraining stage (pSp encoder training), we used the \href{https://github.com/yuval-alaluf/restyle-encoder}{restyle-encoder} implementation, manually adapted for the use with StyleGAN2-ADA generator. ReStyle can be seen as a generalization of pSp with only one cascade step. The hyperparameters:

\begin{table}[h!]
\begin{center}
\resizebox{1.0\linewidth}{!}{
\begin{tabular}{c|c|c|c}
$L_2$ weight $\lambda_1$ & LPIPS weight $\lambda_2$ & ID weight $\lambda_3$ & reg weight $\lambda_4$ \\ \hline
1                        & 0.8                      & 0                     & 0                     
\end{tabular}
}
\end{center}
\end{table}

ID weight $\lambda_3$ was disabled on purpose to avoid the identity information leaking into the encoder during training, so that the experiment is fair. The input images are sampled uniformly from $\mathcal{D}^{prior}$ in $112 \times 112$ resolution. Since the output of the generator is of $128 \times 128$ resolution, we bilinearly downscale the generator output $\hat{I}$ to $112 \times 112$ px before calculating the loss, which is equal to $\| I - \hat{I} \|_2 + 0.8 \cdot \textrm{LPIPS}(I, \hat{I})$. The network follows IR-SE-50 architecture, which is an improved version of ResNet-50 with squeeze-and-excitation modules~\cite{Hu2018squeeze}. The encoder is trained for 16 million 1-sample steps. The training was performed on either 3 or 4 GPUs (either NVIDIA RTX 2080 TI or NVIDIA RTX 3090) with a minibatch of 48 or 64, respectively (depending on the experiment). 

For the final fine-tuning stage (training the network for the face recognition task), we used the \href{https://github.com/ZhaoJ9014/face.evoLVe}{face.evoLVe}~\cite{Wang2021face} library for high-performance face recognition training, which was significantly modified. 
Before training, we copy all weights of the encoder from the first layer through the map2style blocks, excluding the latter, into the backbone, and attach a randomly initialized output block (BatchNorm + Dropout + fully-connected + BatchNorm, as recommended in the implementations (e.g. ~\cite{Wang2021face})). Additionally, we introduce a dropout layer with 0.15 dropout rate after every convolutional layer. The network is trained for 100 epochs. For the first 3 epochs, we freeze all layers except the first convolutional layer and the output block, and after the $3^\mathrm{rd}$ epoch we unfreeze all layers. The optimizer is SGD with momentum of 0.9, weight decay of $2 \cdot 10^{-3}$, and the initial learning rate of 0.03 which is decreased by 1.5 every 5 epochs. Despite the fact that the learning rate setting, its scheduler, the introduction of the dropout layers, and higher weight decay compared to the standard ArcFace pipeline were modified, we found empirically that it helps consistently reproduce the results and better prevent overfitting in a general setting. Augmentations include resizing to $128 \times 128$, random cropping a $112 \times 112$ region, and horizontal flipping with 50\% probability.

The network was trained on 3-5 GPUs (either NVIDIA RTX 2080 TI or NVIDIA RTX 3090) with batch size varying from 300 to 900, depending on the experiment (no significant sensitivity to the batch size parameter in that range was observed). \end{appendix}

\end{document}